\documentclass{article}

\usepackage{arxiv}

\usepackage[utf8]{inputenc} % allow utf-8 input
\usepackage[T1]{fontenc}    % use 8-bit T1 fonts
\usepackage{hyperref}       % hyperlinks
\usepackage{url}            % simple URL typesetting
\usepackage{booktabs}       % professional-quality tables
\usepackage{amsfonts}       % blackboard math symbols
\usepackage{nicefrac}       % compact symbols for 1/2, etc.
\usepackage{microtype}      % microtypography
\usepackage{lipsum}
\usepackage{graphicx}

\usepackage{algorithmic}
\usepackage{textcomp}
\usepackage{stfloats}
\usepackage{verbatim}
\usepackage{subcaption}
\graphicspath{ {./images} }
\usepackage{caption}
\usepackage{amsmath}
\usepackage{tabularx}
\usepackage[normalem]{ulem}
\usepackage{algorithm}

\begin{document}

\title{Empirical Evaluation of Public HateSpeech Datasets} 

\author{
 Sardar Jaf \\
  School of Computer Science \\
  University of Sunderland \\
  Sunderland\\
  SR1 3SD\\
  United Kingdom\\
  \texttt{sardar.jaf@sunderland.ac.uk} \\
   \And
 Basel Barakat \\
  School of Computer Science \\
  University of Sunderland \\
  Sunderland\\
  SR1 3SD\\
  United Kingdom\\
  \texttt{basel.barakatf@sunderland.ac.uk} \\
  }

\maketitle

\begin{abstract}
Despite the extensive communication benefits offered by social media platforms, numerous challenges must be addressed to ensure user safety. One of the most significant risks faced by users on these platforms is targeted hate speech. Social media platforms are widely utilised for generating datasets employed in training and evaluating machine learning algorithms for hate speech detection. However, existing public datasets exhibit numerous limitations, hindering the effective training of these algorithms and leading to inaccurate hate speech classification. This study provides a comprehensive empirical evaluation of several public datasets commonly used in automated hate speech classification. Through rigorous analysis, we present compelling evidence highlighting the limitations of current hate speech datasets. Additionally, we conduct a range of statistical analyses to elucidate the strengths and weaknesses inherent in these datasets. This work aims to advance the development of more accurate and reliable machine learning models for hate speech detection by addressing the dataset limitations identified.
\end{abstract}

%\keywords{Dataset Evaluation \And Hatespeech \and Hate Classification \and Hatespeech Dataset \and Hatespeech Dataset Evaluation \and Hatespeech Corpus Evaluation}

\section{Introduction}
\label{section.Introduction}

Social media is  one of the most widely used online medium for sharing and communication methods for people in modern society, where people can easily share information, news, updates and their opinions on current trends. 
One of the potential risks associated with easily sharing and publishing information by online users that is accessible worldwide is the integrity of the information. Particularly, the dissemination of hatespeech.
Thus, one of the pressing needs for many online platforms and users is to ensure published information on online platforms are free from hatespeech.

To address this need, numerous efforts have been made  for the provision of datasets to be used in the design of AI systems to efficiently and accurately detect, classify and remove hatespeech content. However, despite the availability of many datasets, which are crucial component for the development of AI powered hatespeech detection/classification systems, the quality of those datasets are questionable. Generally, poor quality datasets would lead to poor AI systems. Therefore, in this paper, we focus on evaluating public hatespeech datasets that are based on social media platforms. Our aim is to  empirically evaluate the quality of many public hatespeech datasets in order to assess their suitability for AI hatespeech classifiers.

The contributions of this study are as follows:

\begin{itemize}
    \item We offer the first empirical evaluation attempt of a large number of datasets.
     \item We empirically demonstrate that the quality of dataset content has a greater positive impact on AI hatespeech classification than factors such as content volume, context diversity, and data modalities.
    \item We novel approach utilizes hatespeech dataset features to identify  correlation between each feature and machine learning classification performance. This approach has the potential to be generalised to datasets in other domain.
    \item We  present a simple yet highly effective  baseline deep neural network architecture for hatespeech classification, the outperforms some published binary hatespeech classifiers.
\end{itemize}

The rest of the paper is organised as follows: in section~\ref{section.relatedWork}, we review published literature on hatespeech datasets and machine learning approaches to hatespeech classification. In section~\ref{section.Methodology}, we  describe our methodology, identifying and selecting public datasets, preprocessing them, normalizing them, binarising their labels, performing statistical analyses, and developing a baseline hatespeech classifier. We present experimental setup, results, and detailed analysis of the evaluation results of all the selected datasets in section~\ref{section.evaluation}. 
our experimental setup, results, and detailed analysis of the results. Finally, we conclude our finding and identify potential future work in section~\ref{section.conclusion}.

\section{Literature Review}
\label{section.relatedWork}
One of the primary sources for collecting big data for text analyses is social media platforms. These platforms have been used by researchers from different disciplines as a data collection source~\cite{Soto2018DataQC}. For academic research projects, social media data has been explored widely for research and practical applications of hatespeech detection, analyses and classification. As a result, many datasets have been compiled from various social media platforms for hatespeech processing. In this section, we will highlight core aspects of some of the published datasets and machine learning applications for the  task of hatespeech classification.

\textbf{Hatespeech datasets} are largely produced by extracting content (e.g., text, images, memes, videos, emojis etc.) from  social media platforms, online forums, blogs and various other online communities. 

To develop machine learning approaches to process hatespeech (which involves data analyses, classification, visualisation etc.), access to labelled corpora is essential. Since there is no commonly accepted benchmark corpus for processing hatespeech, authors usually collect from online platforms and annotate them using different annotation approaches. This practice resulted in considerable variation in the size of the published datasets, topics, domains, languages, hatespeech categories, platforms, content types, etc. Some datasets are very large (containing over hundred thousand entries~\cite{Kennedy}~\cite{Gomez} \cite{Djuricetal2015}) whereas others are small (contain a few thousands entries~\cite{Davidson}~\cite{salminen}~\cite{dinaka212}) or few hundreds entries~\cite{Suryawanshi}. The main reasons for such data size variation are: (i) as in any text annotation, annotating hatespeech is an extremely time-consuming process, (ii) there are, usually, much fewer hateful than non-hateful (neutral) comments present in randomly sampled data from social media platforms. Therefore, accomplishing this task necessitates the collection of extensive datasets that can be annotated to identify a substantial number of hatespeech instances. The negative impact of this imbalanced distribution of content types is that it would generally be difficult to build a balanced dataset, where there are equal samples of hateful and neutral content.

Some authors attempted to increase the sample size of hatespeech content whereas keeping the size of data instances to be annotated at a reasonable level, \cite{waseemhovy2016hateful}\footnote{The dataset is available are http://github.com/zeerakw/hatespeech} proposed an approach to pre-select the text instances to be annotated by querying an online platform (Twitter) for topics that are likely to contain a higher degree of hatespeech (e.g. ``Islam terror''). The strength of this approach is it increases the proportion of hatespeech samples in the resulting dataset, and thus resulting in the possibility of achieving a balanced dataset. However, the limitation of this approach is it focuses the resulting dataset on specific topics and certain subtypes of hatespeech (e.g. hatespeech targeting Muslims)\cite{schmidt-wiegand-2017-survey}.

Since there is no commonly accepted benchmark corpus for hatespeech classification, authors usually collect and label their own data \cite{schmidt-wiegand-2017-survey}. For this reason, most of the available datasets are based on content from one or few data sources. Some of the major sources of datasets are: Yahoo\cite{Nobataetal2016} \cite{warner-hirschberg-2012-detecting} \cite{Djuricetal2015}, Twitter\cite{Davidson} \cite{waseemhovy2016hateful}, Reddit \cite{vidgenetal21b}, Qian et al.\cite{Qian}, YouTube\cite{Kennedy}\cite{salminen}, Facebook\cite{salminen}, Stormfront\cite{Gibert} or dynamically generated text\cite{Vidgenetal21a}. The result of collecting data from different online platforms for creating hatespeech dataset is that the dataset are likely to have different characteristics, and subtypes of hatespeech\cite{schmidt-wiegand-2017-survey}. This is largely because of the nature and purpose of the online platforms, and thus they may have special characteristics. For instance, a platform especially created for adolescents, one should expect quite different types of hatespeech than on a  a platform that is used by a cross-section of the general public since the resulting different demographics will have an impact on the topics discussed and the language used \cite{schmidt-wiegand-2017-survey}.

\begin{table}[t!]
\centering
\caption{Dataset names, platforms, year and publication sources.}
   \resizebox{0.5\textwidth}{!}{%
\label{tab:platforms}
\begin{tabular}{@{}llll@{}}\toprule
{Datasets}                 & Platform                              & Year                       & Ref.\\\midrule
Davidson et al.          & Twitter                               &  2017                      & \cite{Davidson}\\
Gibert et al.                   & Stormfront                            & 2018                       & \cite{Gibert}\\
Gomez et al.    & Twitter                               &  2019                      & \cite{Gomez}\\
Kennedy et al.                 & Twitter, Reddit, YouTube              &  2020                      & \cite{Kennedy}\\
Qian et al.             &  Gab                                  &  2019                      & \cite{Qian}\\
Salminen et al.                 & YouTube and Facebook                  & 2018                       & \cite{salminen}\\
Suryawanshi et al.        & Reddit, Facebook, Twitter and Instagram   & 2020                   & \cite{Suryawanshi}\\
Vidgen et al. A                 & Dynamically generated               &  2021                      & \cite{Vidgenetal21a}\\
Vidgen et al. B          & Reddit                                &  2021                      & \cite{vidgenetal21b} \\
Waseem and Havoy         & Twitter                               &  2016                      & \cite{waseemhovy2016hateful} \\ 
\end{tabular}
}
\end{table}

The above issues related to hatespeech datasets have lead to the creation and availability of several datasets for the task of automated hatespeech classification. Table~\ref{tab:platforms} contains a list of publicly available hatespeech datasets, which we have used them in this study. Recent approaches to hatespeech classification usually involves training machine learning algorithm(s) on sample data.

\textbf{Hatespeech classification} methods for processing hatespeech content, especially classifying social media content as ``Hateful'' or ``Neutral'' are largely based on supervised classification method. This method, involves using labelled/annotated data for training machine learning algorithms to classify hatespeech content. Two types of machine learning algorithms are usually used in supervised learning: Shallow learning algorithms (such as support vector machine, decision trees, nearest neighbours etc) have been utilised widely. (ii) Deep learning algorithms, which mainly covers the various types of recurrent neural networks \cite{MehdadandTetreault16}. 

Other classification methods to hatespeech content employ semi-supervised method, particularly bootstrapping, which can be utilized for different purposes in the context of hatespeech processing. On the one hand, it can be used to obtain additional training data, as it is done in~\cite{Xiangetal15}. On the other hand, bootstrapping can  be utilized to build lexical resources used as part of the detection process. The authors, of ~\cite{Gitari2015ALA}, apply this method to populate their hate verb lexicon, starting with a small seed verb list, and iteratively  expanding it based on WordNet relations, adding all synonyms and hypernyms of those seed verbs.

\section{Methodology}
\label{section.Methodology}
\subsection{Data collection}
Hatespeech is gaining increasing attention from industry, government organisations and academia. The proliferation of information published on social media platforms provides the means to create datasets for processing, analysing, detecting and classifying hatespeech content. Variations in the available datasets (e.g, in size, topic, domain, language, hatespeech categories, platform, content type, etc.) could be beneficial. However, it can make it challenging for researchers to determine which dataset is suitable for training machine learning algorithms for hatespeech classification. For example, a dataset based only on Twitter content may, or may not, be suitable for producing a generalisable machine learning model that perform well on non-twitter data, such as YouTube comments.

The aim of this study is to evaluate multiple publicly available dataset to assess their suitability for training and testing deep learning algorithm for hatespeech classification. We have selected ten datasets from hatespeechdata.com website, which is a widely used platform for hosting hatespeech dataset. Our goal in evaluating multiple dataset is to examine two application aspects of each dataset: (i) to examine the suitability of a dataset in testing the performance of a deep learning based system for hatespeech classification, and (ii) to examine the suitability of a dataset in producing generalisable deep learning model by training a deep learning algorithm on a dataset but testing it on other dataset with different domain, text, genre, and size. The selected dataset names, platforms where the data are collected from, and the publication years of the dataset are presented in Table~\ref{tab:platforms}. We chose the evaluated datasets based on several characteristics: different platforms, dataset size, content type, length of individual text entry, and publication time. Examining dataset with content extracted from various platforms helps us to evaluate the generalisability of deep learning algorithms. Similarly, the different years were chosen to ensure that the evaluation would be generalisable, as some hateful terms might be more popular in specific years. We selected datasets with different sizes because machine learning algorithm performance is usually dependent on dataset size. 

The labelled content in each selected dataset varies from one dataset to another. The proposed dataset by~\cite{vidgenetal21b}, has various categories of abuse (e.g., targeted Identity, affiliation, and person), and counter speech. The published dataset by~\cite{Davidson} has three types of content (classes) i.e., \textit{Hatespeech}, \textit{Offensive language} and  \textit{Neither}.
The datasets, from~\cite{Qian} and~\cite{Vidgenetal21a}, have only two classes: \textit{Hate}/\textit{Not Hate} and \textit{Offensive}/\textit{Vulgar}.
Kennedy et al.'s dataset~\cite{Kennedy} content is focused on particular categories (such as religions,  three hate classes for races , and a hate score  to indicate the hate level).
The dataset from \cite{Gomez} contains text and images. They are labelled as ``Hate'' or ``Not-Hate''.``Hate'' content is further divided into five classes of different types of hate.  Waseem and Havoy's dataset~\cite{waseemhovy2016hateful}, has three labels: ``Sexism'', ``Racism'', and ``None''. 

The content of the available dataset has been labelled with different types of hatespeech. Therefore, there are inconsistent labels between the dataset. In order to design and evaluate a supervised deep learning hatespeech classifier trained on the available dataset for binary hatespeech classification, we have converted the various labels in the dataset into either ``Hate'' or ``Not-Hate'' labels. Therefore, each dataset  has one of two classes (``Hate'' or ``Not-Hate''). This  approach enables us to make the labels in all the selected datasets uniform. 

\begin{table}[t]
\centering
\caption{Dataset and examples of content categories}
 \resizebox{0.7\textwidth}{!}{%
\begin{tabular}{l p{0.75\linewidth}} 
\toprule
\textbf{Datasets}        & {Examples content categories}	     \\  \midrule   
  Davidson et al.~\cite{Davidson} & hatespeech, offensive language, neither \\ 
  Gibert et al.~\cite{Gibert}     &     Hate/not hate,relation, idk/skip  \\
  Gomez et al.~\cite{Gomez}       &    Hate/not hate         \\ 
  Qian et al.~\cite{Qian}         &   Hate, Offensive/Vulgarity    \\ 
  Kennedy et al.~\cite{Kennedy}    &   respect, insult, humiliate, status, dehumanize, violence, genocide, attack\_defend, target\_race\_... (black, Asian, native American, white, middle eastern etc), target\_religion\_... (e.g., atheist, buddhist, christian, hindu, jewish etc), target\_origin\_... (e.g. immigrant, migrant\_worker, specific\_country etc) target\_gender\_... (e.g. men, women, transgender\_men, etc), target\_sexuality\_... (e.g., bisexual, gay, lesbian, straight, etc), target\_age\_... (e.g., children, teenagers, young\_adults, etc), target\_disability\_... (e.g., physical, cognitive, neurological, visually\_impaired, etc.)     \\ 
  Salminen et al.~\cite{salminen}       &     Hate/       Neutral  \\
  Suryawanshi et al.~\cite{Suryawanshi}  &     Offensive/Non-offensive  \\
  Vidgen et al. A~\cite{Vidgenetal21a}   &    Hate/not hate   \\
  Vidgen et al. B~\cite{vidgenetal21b}   &  AffiliationDirectedAbuse, PersonDirectedAbuse, IdentityDirectedAbuse, CounterSpeech  \\ 
  Waseem and Havoy~\cite{waseemhovy2016hateful}   &    Sexism, racism \\     
\end{tabular}
}
\label{tab:dataset.hate.categories}
\end{table}

\subsection{Label binarization} \label{subsub:label.binarazation}
The available datasets have multiple types (classes) of hatespeech content. Each dataset contains different types of hatespeech, which results in inconsistent hatespeech labels between the datasets. One of the main reasons that published hatespeech datasets have different labels for hatespeech content is because there is no uniform consensus in the research community on the different  classes of hatespeech, which is a challenge that requires further effort from the research community to address. Since there is no  consensus in the research community on the different types of hatespeech, authors of the published datasets have labelled their data with different classes of hate. Some datasets contain fine-grained categories of hate, where the victims are targeted based on their race; religion; sexuality; ethnicity; gender etc.~\cite{Kennedy, vidgenetal21b}. Other datasets contain broad categories  such as ``hateful'', ``abusive'' or ``neutral'' \cite{Gomez,Vidgenetal21a, waseemhovy2016hateful}.
 Table~\ref{tab:dataset.hate.categories}  presents a summary of some of the categories of hatespeech in each dataset. Some datasets  contain very few and general hatespeech content  (such as ``hate''  or ``offensive'', as in  the dataset published by~\cite{Qian} and ~\cite{Davidson}) whereas other datasets have many and fine-grained hatespeech types such as the dataset published by~\cite{Kennedy} which includes various subtypes of hatespeech based on gender or religion.

Prior to exploring and analysing the content of the selected datasets for this study, we have binarised the labels. The objective is to ensure all the datasets contain consistent labels (``Hate'' or ``Not-Hate''). The label binarization process is performed as follows:

\begin{itemize}
    \item Merge fine-grained labels of hatespeech to broad labels. If the label of content indicates any type of hatespeech, then we convert it to a broad label ``Hate''. If the label indicates the content is ``neutral'' or ``not hateful'', then we convert to ``Not-hate''.
    \item Drop ambiguous labels by discarding any content in a dataset where the content has an ambiguous label, such as ``abusive'', because such content may not be considered hateful.
    %This approach is consistent with the work of \textbf{[REF]}.
    \item We convert content that is labelled as ``neutral'', ``not-hate'', or ``not-abusive''  to the``Not-hate'' label.
\end{itemize}

\begin{table}[t]
\centering
    \caption{Datasets statistics}
      \resizebox{0.6\textwidth}{!}{%

    \begin{tabular}{llllll}
     \begin{tabular}[c]{@{}l@{}}Datasets \\     Name\end{tabular}               & \begin{tabular}[c]{@{}l@{}}Split     \\ (Hate / Not-Hate) \\Hate \% \end{tabular} & & \begin{tabular}[c]{@{}l@{}}Words Count \\ Mean/Median    \\ (Hate) \\ Not-Hate\end{tabular} & \begin{tabular}[c]{@{}l@{}} Words Count \\  Min/Max    \\ (Hate) \\ Not-Hate\end{tabular} & \begin{tabular}[c]{@{}l@{}} Words Count \\ VAR/STD    \\ (Hate) \\ Not-Hate\end{tabular}   \\ \midrule
 Davidson et al.~\cite{Davidson}                   & \begin{tabular}[c]{@{}l@{}}(1430 / 1430)\\ 50\%\end{tabular}         &                       & \begin{tabular}[c]{@{}l@{}}(13.9/ 13.0)\\ 14.8/15\end{tabular}        & \begin{tabular}[c]{@{}l@{}}(1/ 32)\\ 2/32\end{tabular}          & \begin{tabular}[c]{@{}l@{}}(49.2/7.0)\\ 45.5/6.7\end{tabular}       \\\midrule
Gibert et al.~\cite{Gibert}         & \begin{tabular}[c]{@{}l@{}} (1437/ 9507) \\ 13.1\%\end{tabular} & &\begin{tabular}[c]{@{}l@{}}(22.0/20.0)\\ 17.3/15.0\end{tabular} &\begin{tabular}[c]{@{}l@{}}(1/349)\\ 1/262\end{tabular} &\begin{tabular}[c]{@{}l@{}}(234.8/15.3)\\ 179.5/13.4\end{tabular}  \\ \midrule  
    Gomez et al.~\cite{Gomez}                   & \begin{tabular}[c]{@{}l@{}}(112787 / 25263) \\ 81.7\%\end{tabular}     &                        & \begin{tabular}[c]{@{}l@{}}(11.7/11.0)\\ 11.5/11\end{tabular}         & \begin{tabular}[c]{@{}l@{}}(2/90)\\ 2/81\end{tabular}             & \begin{tabular}[c]{@{}l@{}}(28.4/5.3)\\ 28.4/5.3\end{tabular}       \\\midrule
    Kennedy et al.~\cite{Kennedy}                    & \begin{tabular}[c]{@{}l@{}}(46021 / 80624)\\ 36.3\%)\end{tabular}      &                        & \begin{tabular}[c]{@{}l@{}}(25.8/19.0)\\ 28.1/21\end{tabular}         & \begin{tabular}[c]{@{}l@{}}(1/128)\\ 1/128\end{tabular}           & \begin{tabular}[c]{@{}l@{}}(412.4/20.3)\\ 507.0/22.5\end{tabular}   \\\midrule
    Qian et al.~\cite{Qian}                        & \begin{tabular}[c]{@{}l@{}}(2348 / 25198)\\ 8.5\%\end{tabular}        &                       & \begin{tabular}[c]{@{}l@{}}(27.8/23.0)\\ 20.3/15\end{tabular}         & \begin{tabular}[c]{@{}l@{}}(1/191)\\ 1/282\end{tabular}           & \begin{tabular}[c]{@{}l@{}}(364.5/ 19.1)\\ 274.1/16.6\end{tabular}  \\\midrule
    Salminen et al~\cite{salminen}   & \begin{tabular}[c]{@{}l@{}}(2364 / 858) \\ 73.4\%\end{tabular}     & & \begin{tabular}[c]{@{}l@{}}(43.6/34.0)\\ 38.2/30.0\end{tabular} &\begin{tabular}[c]{@{}l@{}}(1/386)\\ 1/351\end{tabular} &\begin{tabular}[c]{@{}l@{}}(1678.1/41.0)\\ 1408.5/37.5\end{tabular} \\ \midrule
    Suryawanshi et al.   & \begin{tabular}[c]{@{}l@{}}(303 / 440) \\ 40.8\%\end{tabular}      & & \begin{tabular}[c]{@{}l@{}}(45.0/33.0)\\ 44.8/32.0\end{tabular} &\begin{tabular}[c]{@{}l@{}}(4/307)\\ 2/268\end{tabular} &\begin{tabular}[c]{@{}l@{}}(1630.4/40.4)\\ 1743.2/41.8\end{tabular}  \\\midrule
    Vidgen et al. A~\cite{Vidgenetal21a}                    & \begin{tabular}[c]{@{}l@{}}(22175 / 18969) \\ 53.9\%\end{tabular}      &                        & \begin{tabular}[c]{@{}l@{}}(23.8/15.0)\\ 25.1/17\end{tabular}         & \begin{tabular}[c]{@{}l@{}}(1/395)\\ 1/408\end{tabular}           & \begin{tabular}[c]{@{}l@{}}(599.6/24.5)\\ 621.0/24.9\end{tabular}   \\\midrule
     Vidgen et al. B~\cite{vidgenetal21b}             & \begin{tabular}[c]{@{}l@{}}(4093 / 19107)\\ 17.6\%\end{tabular}        &                       & \begin{tabular}[c]{@{}l@{}}(39.5/ 19.0)\\ 28.7/14.0\end{tabular}      & \begin{tabular}[c]{@{}l@{}}(1/1937)\\ 1/1417\end{tabular}         & \begin{tabular}[c]{@{}l@{}}(7690.7/87.7)\\ 2908.0/53.9\end{tabular} \\\midrule 
     Waseem and Havoy~\cite{waseemhovy2016hateful}                     & \begin{tabular}[c]{@{}l@{}}(2692 / 7766)\\ 25.7\%\end{tabular}        &                        & \begin{tabular}[c]{@{}l@{}}(16.8/17)\\ 14/14\end{tabular}             & \begin{tabular}[c]{@{}l@{}}(1/33)\\ 1/38\end{tabular}             & \begin{tabular}[c]{@{}l@{}}(41.3 /6.4)\\ 49.3/7.0\end{tabular}      \\ \midrule                                           
    
    \end{tabular}
    }
    \label{tab:label.binarization}
    \end{table}

Table~\ref{tab:label.binarization} presents some statistical information on each dataset after we have binarised the labels as ``Hate'' or ``Not-Hate''.  The table contains the followings: sample size for each dataset based on the content split between ``Hate'' and ``Not-Hate'', total unique word count for ``Hate'' and ``Not-Hate'' content, the mean and median, the min/max and the variance and standard deviation.

The label binarization process provides us with a dataset containing consistent labels of ``Hate'' or ``Not-Hate'',  which we can use to  evaluate their suitability for training and testing a deep learning algorithm for binary hatespeech classification. We evaluate a baseline deep learning system on each dataset to assess its performance in two tasks: (i) performing binary classification of hatespeech (i.e., classifying text as either ``Hate'' or ``Not-hate''), and (ii)  performing transfer learning classification to test the generalisation of the system where we train the system on one dataset and test it on multiple other dataset. 

One of the major issues with all the public datasets that negatively impact the performance of machine learning algorithms is the imbalance in the features. The sample size for different categories of text is often uneven, with some categories having more content than others. Most of the datasets appear to have more text related to ``Not-Hate''  categories than ``Hate''. To address the data imbalance problem, we balance the sample size of ``Hate'' and ``Not-Hate'' content for each dataset before we use them  to design and evaluate a baseline deep learning system.

\begin{table}[t]
\centering
\caption{Dataset size: before and after balancing sample size}
 \resizebox{0.6\textwidth}{!}{%
\begin{tabular}{lllllll} & \multicolumn{3}{c}{\textbf{Initial Binarized Dataset Size}} & \multicolumn{3}{c}{\textbf{Balanced Binarized Dataset Size}} \\
\textbf{Dataset}         & Hate           & Not-Hate         & Total          &  Hate           & Not-Hate         & Total    \\
\midrule
Davidsonet al~\cite{Davidson}                  & 1430           & 1430             & 2860           & 1430             & 1430             & 2860            \\
Gibert et al.~\cite{Gibert}                   & 1437           & 9507             & 10944          & 1437             & 1437             & 2874 \\ 
Gomez et al.~\cite{Gomez}                 & 25263          & 112787           & 138050         & 25263            & 25263            & 50526           \\
Kennedy et al.~\cite{Kennedy}                 & 46021          & 80624            & 126645         & 46021            & 46021            & 92042           \\
Qian et al.~\cite{Qian}                      & 2348           & 25198            & 27546          & 2348             & 2348             & 4696            \\
Salminen et al~\cite{salminen}                & 2364           & 858              & 3222           & 858              & 858              & 1716            \\
Suryawanshi et al.~\cite{Suryawanshi}                & 303            & 440              & 743            & 303              & 303              & 606             \\
Vidgen et al. A~\cite{Vidgenetal21a}                   & 22175          & 18969            & 41144          & 18969            & 18969            & 37938      \\
Vidgen et al. B~\cite{vidgenetal21b}           & 4093           & 19107            & 23200          & 4093             & 4093             & 8186            \\
Waseem and Havoy~\cite{waseemhovy2016hateful}                   & 2692           & 7766             & 10458          & 2692             & 2692             & 5384 \\
\end{tabular}
}
\label{tab:balanced.dataset.size}
\end{table}

\subsection{Dataset balancing}
\label{subsub:data.balancing}
The available datasets are imbalanced, as they contain unequal sample size for different text categories. As we discussed in Section~\ref{subsub:label.binarazation}, we have binarised the labelled content in each dataset so that they contain only ``Hate'' and ``Not-Hate''. 

As it can be seen in Table~\ref{tab:balanced.dataset.size}, the differences in the sample size for different labelled content is large in all the datasets. Such differences in sample sizes negatively affect the training of machine learning algorithms (including deep learning algorithms), as the algorithms become biased towards the majority sample. To balance the sample size of ``Hate'' and ``Not-Hate'' content in each dataset, we apply under-sampling methods, reducing the size of the majority sample to match the size of the minority one. The columns in Table~\ref{tab:balanced.dataset.size} show the balanced sample size of ``Hate''/``Not-Hate'' content for each dataset.

\subsection{Statistical Analysis}
\label{subsec.statistical.analysis}

To gain a better understanding of the nature and prevalence of hatespeech, this study utilised a quantitative approach to analyse the frequency of  hate terms in both hateful and non-hateful speech. Our analysis began with collecting, 1523 frequently used hate terms from Hatebase\footnote{Available at: https://hatebase.org/}, a publicly available database of hatespeech terms. We then conducted an analysis of the usage frequency of these terms in both types of speech. To ensure the comparability of the two counts, we utilised the balanced datasets.

\begin{figure}[t]
    \centering
    \includegraphics[width=0.5\columnwidth]{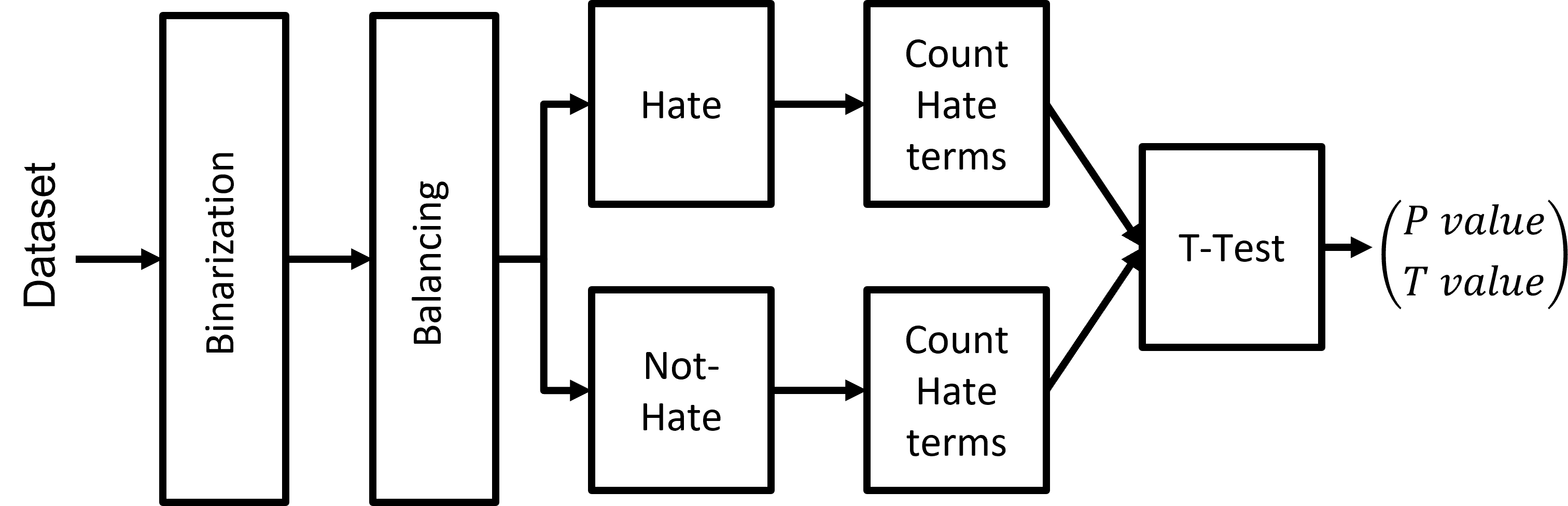}
    \caption{Block diagram of the T-test procedure used to compare the usage of hate terms in hateful and non-hateful speech in each dataset}
    \label{fig:t-test-block}
\end{figure}

We conducted a T-test for the means of the counted hate terms frequency using python  scientific computing package~\cite{scipy}. The T-test generated two matrices for the comparison, the $t$-value and the $p$-value, providing a quantitative measure of the significance of differences in the usage of hate terms between hateful and non-hateful speech. This allowed us to compare the usage of the hate terms in both types of speech within each dataset, and to draw conclusions about the prevalence and nature of hatespeech in our datasets. Fig \ref{fig:t-test-block} presents a block diagram of the used T-test procedure.

\subsection{Hatespeech Classification}
\label{sub.text.classification.prep}
\textbf{Text preprocessing: cleaning and normalization.}
\label{subsub:preprocesingClassification}
Since the content of the published dataset is collected from different online platforms (e.g., Twitter, Facebook, YouTube comments, etc.), they have different features such as structure, topic, user writing style, etc. We stored each dataset in comma separated value files (CSV) with two columns (text and label). The text column contains the text content and the label column contains one of two values, ``hate'' or ``not-hate''. We have performed the following transformations on the text before training and evaluating our model:
\begin{itemize}
    \item Lower casing. We convert all the text to lower case English characters.
    \item Removing non-English text. We remove content that is not part of the English alphabet.
    \item Normalising emojis by replacing with token ``$<$EMOJ$>$''.
    \item Normalising tag. We transform all hashtags to the token ``$<$HASHTAG$>$'' and all usernames to the token ``@USER''.
    \item Removing duplication. We remove sequentially duplicated items such as words, spaces, characters etc.
    \item Removing punctuation. We remove all the English and non-English punctuation.
    \item Removing stop words. We remove all stop words such as `a', `the', `of' etc.
    \item Normalizing URL. We transform all hyperlinks and website address to the token ``$<$URL$>$''.
    \item Normalising HTML elements. We convert all named and numeric character references from HTMLs such as ``\&gt;'' and ``\&\#amps;'' in the text to their corresponding Unicode characters ``$<$'' and ``\&'', respectively.
    \item Removing new line in text. We remove all new line in each text to create a single line text.    
\end{itemize}

\textbf{Deep learning model implementation.}
\label{subsec:deepLearningModel}
We have implemented a baseline deep learning text classification system. We have trained and tested the system on ten publicly available datasets.

Our model is based on a widely utilised deep learning algorithms for text classification. We have also utilised a general-purpose language model that has contributed to many natural language processing tasks. 

After preprocessing and normalizing the text in each dataset~\ref{sub.text.classification.prep}) we have tokenised the input data (textual information) using  Bidirectional Encoder Representations from Transformers model (BERT) \cite{bert}, which is an essential step for training any deep learning algorithms. The BERT model is a multilingual language model trained on a very large text data. BERT model is a multi-layer bidirectional Transformer, which is a deep learning model used in several Natural Language Processing tasks ~\cite{devlinetal19}. 

Next, the vector representation of the data is processed by a dropout layer with a dropout rate of 0.3. This step involves randomly excluding 30\% of the training data during the training phase of the model in order to prevent the algorithm from memorizing the data pattern from the dataset, as is usually referred to as overfitting. 

\begin{figure}[t]
    \centering
    \includegraphics[trim={1cm 6cm 1cm 1cm},clip, width=0.7\columnwidth]{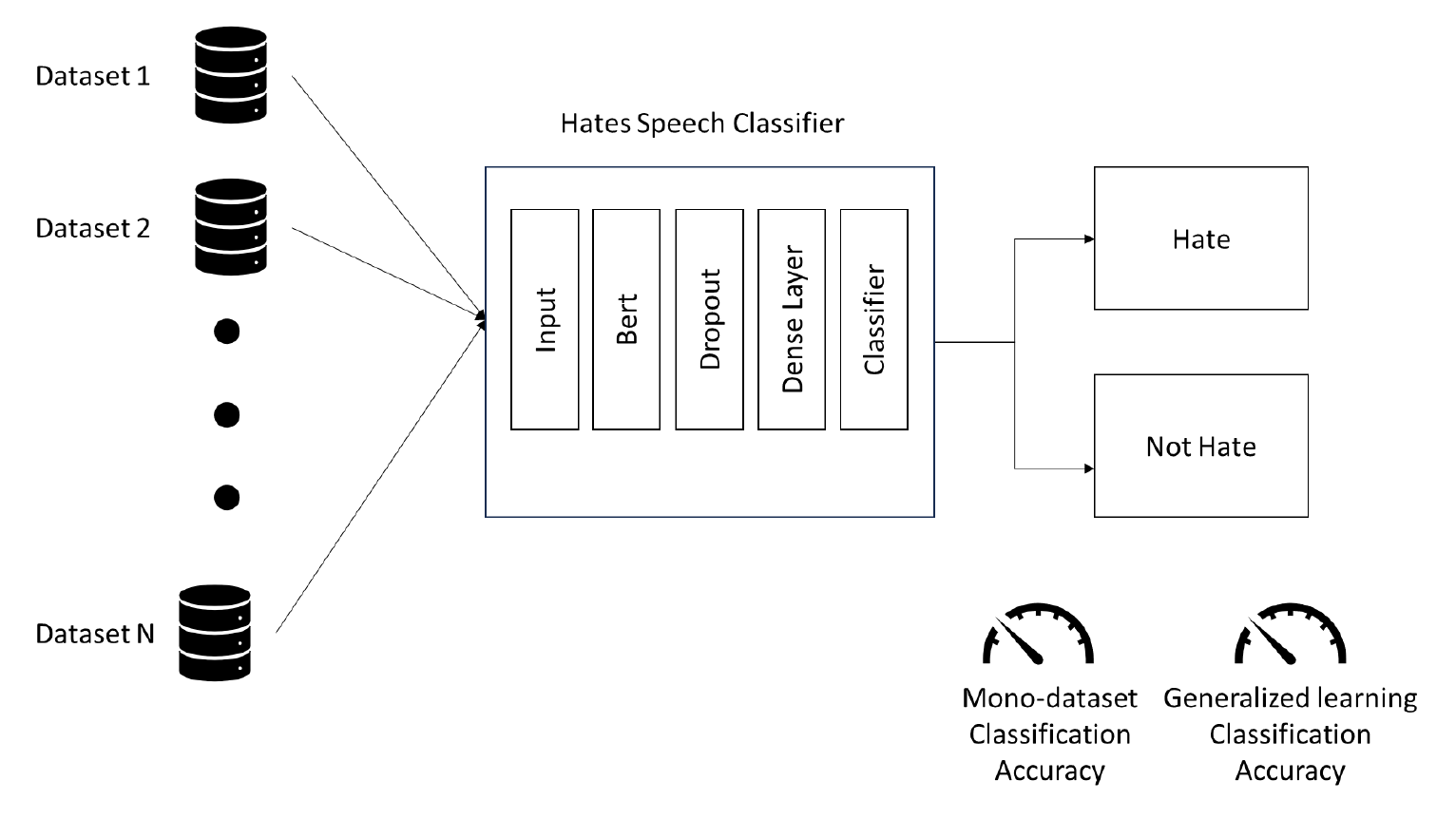}
    \caption{Baseline architecture}
    \label{fig:model.architecture}
    %\Description{deep learning architecture}
\end{figure}

The output from the dropout later is used as input to a single deep learning dense layer of neural network. We optimised the model learning capacity using Adam optimizer with learning rate of $2e-5$. During the training phase, we computed the training error rate using a Binary Cross Entropy Loss function. Fig~\ref{fig:model.architecture} shows the components of our model architecture.

BERT assigns special tokens to sentences, the ``CLS'', ``SEP'' and ``PAD'' token. The ``CLS'' token is to indicate for the model to identify the start of a sentence. The ``SEP'' token to indicate the end of a sentence for the model. Therefore, using ``CLS'' and ``SEP'' informs the model of the start and end of a sentence. We use the ``PAD'' token is used to assign extra empty token(s) to sentences in order for all input sentences to have equal token length. For example, if we restrict the model to learn from sentences of 50 tokens long, and  a sentence contains 30 tokens we would be appended to it to 20 empty tokens (``PAD'') to make it a 50 tokens long sentence. We set the maximum length of sentences to 128 tokens. We add padding token ``PAD'' to sentences with less than 128 tokens in order to reach 128 tokens. Any sentence over 128 tokens long will be truncated. Moreover, BERT employs attention mask where tokens that represent words are masked with the value of 1 and tokens that represent nothing (e.g., ``PAD'') are masked with the value of 0. This masking mechanism allows BERT to determine what token(s) to retain or discard during the learning phase of the model. Each sentence, after being separated into individual words, will be encoded into vectors. We use the bert-base-cased pre-trained version for our model. BERT has been trained on 110 million parameters. The main advantage of using a pre-trained model (such as BERT in our study) is the significant reduction in training time. 

\textbf{Deep learning model Evaluation.}
We follow a standard approach for evaluating a machine learning algorithm for a supervised binary classification task. The evaluation metrics used in this study are based on three standard measures for classification task: Recall, Precision and Weighted F1-score.

\section{Dataset evaluation and results}
\label{section.evaluation}

\subsection{Dataset evaluation}
We  empirically evaluated ten different datasets using a baseline classifier. For each dataset evaluation, we divided the dataset into two parts: a training set and a test set, ensuring  that the content of the two sets was distinct. We allocated 80\% of each dataset to the training set for training the baseline classifier and 20\%  to the test set for evaluating the classifier's performance. within the training set, we further partitioned the data by allocating 90\% for training the deep learning architecture and reserving 10\% for validating the training accuracy during the training stage. 

We conducted  two types of evaluations on each dataset: 
\begin{itemize}
    \item \textbf{Mono-dataset evaluation}. In the model evaluation process, we trained and tested our baseline classifier, individually on each dataset, to classify their content as either ``Hate'' or ``Not-Hate''. The aim of this experiment is to assess the suitability of each dataset for  binary classification of hatespeech content. 
    
    \item \textbf{Generalised learning evaluation}. In this evaluation, we used the baseline classifier that we used for the mono-dataset evaluation, but we tested it differently. In this evaluation, we have conducted multiple experiments on the baseline classifier. In each experiment, we trained the model on one dataset and tested it on nine other datasets. For this experiment, our approach is: given dataset $d$ is a member of a set of dataset $D^{i,...,n}$, we trained our model on $d^i$ and tested it on all the dataset in $D$ except $d^i$. We trained the baseline classifier on the train set of dataset $d^i$ and tested it on the test set of the dataset $D^{j...n}$, where  $d^j$ is the dataset in $D$ that is not the same as the test set of the  dataset $d^i$, as demonstrated in Algorithm\ref{alg.generlised}. This approach is similar to the transfer learning approach, however, the classifiers here are not fine tunned on the test sets. 
    
    This  evaluation method  allows us to assess the dataset's suitability for training a deep learning model that can be effectively generalised to other datasets, which may exhibit different features compared to the one used for training. These variations may include differences in data content published on diverse online platforms, encompassing variations in users' writing styles, content topics, dataset sizes, and other relevant characteristics. We utilise this evaluation to determine each dataset's suitability for producing a deep learning model capable of generalising to classifying ``Hate'' and ``Not-Hate'' content published on different online platforms.
\end{itemize} 

\begin{algorithm}[t]
\caption{Generalised learning evaluation}
\label{alg.generlised}
\begin{algorithmic}[1]
\REQUIRE Dataset set $D = \{d^1, d^2, \ldots, d^n\}$
\FOR{$i = 1$ to $n$}
    \STATE Train the model on $d^i$
    \FOR{$j = 1$ to $n$}
        \STATE Test the model on $d^j \in D$ and $j \neq i$
    \ENDFOR
\ENDFOR
\end{algorithmic}
\end{algorithm}

\begin{table}[t]
\centering
\caption{System performance - mono-dataset classifier}
 \resizebox{0.45\textwidth}{!}{
\begin{tabular}{@{}lllll@{}}
\toprule
\textbf{Model}              &  Weighted F1   & Recall   & Precision   \\ \midrule
Gomez et al.~\cite{Gomez}                   & 0.697         & 0.694 & 0.712     \\
Vidgen et al. B~\cite{vidgenetal21b}        & 0.740         & 0.739 & 0.745     \\
Gibert et al.~\cite{Gibert}                 & 0.777         & 0.774 & 0.803     \\ 
Vidgen et al.~\cite{Vidgenetal21a}          & 0.789         & 0.789 & 0.790     \\
Qian et al.  \cite{Qian}                    & 0.817         & 0.817 & 0.817      \\
Kennedy et al.~\cite{Kennedy}               & 0.841         & 0.840 & 0.841     \\
Waseem and Havoy~\cite{waseemhovy2016hateful}              & 0.879         & 0.879 & 0.882     \\ 
Salminen et al.~\cite{salminen}             & 0.884         & 0.884 & 0.885     \\
Suryawanshi et al. ~\cite{Suryawanshi}      & 0.902         & 0.902 & 0.910    \\
Davidson et al~\cite{Davidson}              & \textbf{0.930} & \textbf{0.930}  & \textbf{0.930}    \\  
\bottomrule
\end{tabular}
}
\label{tab:result.binary.performance}
\end{table}

\subsection{Results}
\label{subsection.result}
We have conducted multiple empirical evaluations of our baseline classifier, which we described in section \ref{subsec:deepLearningModel}, on ten publicly available datasets. In this section, we report the empirical evaluation outcomes of the effectiveness of each dataset for training and testing a baseline deep learning classifier to examine the suitability of different public dataset for hatespeech classification.

For each dataset evaluation, we conducted two experiments: (i) we evaluated the suitability of each dataset for training and testing a baseline classifier for the binary classification of hatespeech content. The training and testing samples are from the same dataset. Thus, we refer to this experiment as ``mono-dataset experiment'', and we refer to the baseline classifier in this experiment as ``mono-dataset classifier''. (ii) The second experiment involved testing the suitability of each dataset to produce a generalised baseline classifier, which  has the same architecture as the mono-dataset classifier but trained and tested in a more generalisable approach. We trained the baseline classifier on a dataset and tested it on nine other datasets, excluding the dataset that we used for training the classifier. We refer to this experiment as ``generalised learning experiment'', and we refer to the baseline classifier  as ``generalised classifier''. 

For each experiment--due to space limitation-- we report the system performance using weighted F1-score, which is based on the harmonic mean of recall and precision.

\subsection{Mono-Dataset Experiment.}
Table \ref{tab:result.binary.performance} presents the performance of the mono-dataset classifier. Out of the ten selected dataset,the model performs best when trained on the dataset published by Davidson et al.~\cite{Davidson} achieving a weighted F1-score of 0.930. The second-best performance is based on the Suryawanshi et al.~\cite{Suryawanshi} dataset with a weighted F1-score of 0.902, which is slightly  behind Davidson et al.~\cite{Davidson}'s. The classifier performed worse when trained and tested on the dataset published by~\cite{Gomez}, producing a weighted F1-score of 0.697.

The classifier achieved moderate performance (between 0.81 and 0.88 of weighted F1-score) when evaluated on the following datasets: Qian et al.~\cite{Qian} (0.817), Kennedy et al.~\cite{Kennedy} (0.841), Waseem and Havoy~\cite{waseemhovy2016hateful} (0.879) and Salminen et al.~\cite{salminen} (0.884). Furthermore, the classifier produced a weighted F1-score between 0.71 and 0.78 when trained and tested on the following datasets: Vidgen et al.~\cite{vidgenetal21b} (0.740), Gibert et al.~\cite{Gibert} (0.777) and Vidgen et al. A~\cite{Vidgenetal21a} (0.789).

Table~\ref{tab:result.binary.rank}, shows the ranking of different datasets in ascending order based on the classifier's performance, measured by weighted F1-score.

\begin{table}[t]
\centering
\caption{System performance - mono-dataset classifier performance rank based on weighted F1 score}
 \resizebox{0.45\textwidth}{!}{
\begin{tabular}{lll}
     \toprule
        \textbf{Model}              & Weighted F1-score &  \\  \midrule
        Davidson et al.~\cite{Davidson}                   & 0.930   & 1 \\ 
        Suryawanshi et al.~\cite{Suryawanshi}                    & 0.902   & 2 \\
        Salminen et al.~\cite{salminen}                   & 0.884   & 3 \\
        Waseem and Havoy~\cite{waseemhovy2016hateful}                      & 0.879   & 4 \\ %\hline
        Kennedy et al.~\cite{Kennedy}                     & 0.840   & 5 \\ 
        Qian et al.~\cite{Qian}                             & 0.816   & 6 \\
        Vidgen et al A.~\cite{Vidgenetal21a}                      & 0.789   & 7 \\ 
        Gibert et al.~\cite{Gibert}                        & 0.777   & 8 \\ 
         Vidgen et al. B~\cite{vidgenetal21b}        & 0.740   & 9 \\ 
        Gomez et al.~\cite{Gomez}              & 0.697   & 10 \\ 
\bottomrule
\end{tabular}
}
\label{tab:result.binary.rank}
\end{table}

The dataset from Gomez et al.~\cite{Gomez}  is the least effective for training and testing a baseline neural network classifier, despite being a large dataset, which, theoretically, should be  beneficial for deep neural network algorithm training. Davidson et al.~\cite{Davidson}'s dataset is ranked first for training a baseline classifier, with Suryawanshi et al.~\cite{Suryawanshi}'s dataset (ranked second)  narrowly behind. Several other datasets  performed moderately: Salminen et al.~\cite{salminen} (0.884), Waseem and Havoy~\cite{waseemhovy2016hateful}  (0.879), Kennedy et al.~\cite{Kennedy}  (0.840) and Qian et al. (0.816). The dataset from Vidgen et al. A~\cite{Vidgenetal21a}, Gibert et al.~\cite{Gibert}, and Vidgen et al.~\cite{vidgenetal21b} produced weight F1-score of 0.789, 0.777 and 0.740, respectively. 

In comparison with the reported hatespeech classifiers proposed by some of the authors of the published datasets, Table~\ref{tab:result.binaryc.comparison} presents the recall, precision, and f1-score of our baseline system and other published binary hatespeech classifiers. The table demonstrates that in some cases, our baseline system trained on the binary labels of the selected datasets outperforms the binary systems proposed by the dataset authors.

\begin{table}[t]
\centering
\caption{System performance comparison - mono-dataset classifier performance against published works.}

 \resizebox{0.8\textwidth}{!}{
\begin{tabular}{lllllll}
     \toprule
        \textbf{Datasets} & \multicolumn{3}{c}{Our baseline system} & \multicolumn{3}{c}{Published systems by dataset authors} \\  
        &   Recall   & Precision & Weighted F1   & Recall   & Precision  & Weighted F1 \\    \midrule
        Davidson et al.~\cite{Davidson}                 & 0.930 & 0.930 & 0.930   & 0.90 & 0.91 & 0.90 \\
        Salminen et al.~\cite{salminen}                  & 0.884 & 0.885 & 0.884   & \_ & \_ & 0.96 \\
        Waseem and Havoy~\cite{waseemhovy2016hateful}    & 0.879 & 0.882 & 0.879   & 0.729 & 0.774 & 0.739 \\
        Qian et al.~\cite{Qian}                         & 0.817 & 0.817 & 0.816   & \_ & \_ & 0.896 \\ 
\bottomrule
\end{tabular}
}
\label{tab:result.binaryc.comparison}
\end{table}

\begin{table*}[t]
\centering
   \caption{System performance: generalised classifier performance based on weighted F1-score}
   \label{tab:generlized.model}
   \resizebox{\textwidth}{!}{
   
    \begin{tabular}{llllllllllll}
     \toprule
\textbf{Model\textbackslash{}Dataset }& Davidson et al. & Waseem and Havoy &  Vidgen et al. B & Salmenin et al. & Gomez et al. & Kennedy et al. & Vidgen et al. A & Suryawanshi et al. & Qian et al.   & Gibert et al. & Mean \\
\midrule
Gomez et al.~\cite{Gomez}                    & 0.292   & 0.373  & 0.363            & 0.341     & *         & 0.395    & 0.425  & 0.486      & 0.253 & 0.341  & 0.363 \\
Suryawanshi et al.~\cite{Suryawanshi}                    & 0.506   & 0.625  & 0.561            & 0.602     & 0.481     & 0.580    & 0.446  & *          & 0.556 & 0.425  & 0.531 \\
Waseem and Havoy~\cite{waseemhovy2016hateful}                      & 0.618   & *      & 0.509            & 0.463     & 0.459     & 0.519    & 0.505  & \textbf{0.722}      & 0.530 & 0.467  & 0.532 \\
Kennedy et al.~\cite{Kennedy}                  & 0.803       & 0.427  & 0.574            & 0.726     & 0.392     & *        & 0.523  & 0.468      & 0.662 & 0.655  & 0.581 \\
Davidson et al.~\cite{Davidson}                     & *       & 0.435  & 0.586            & 0.753     & 0.410     & \textbf{0.729}    & 0.519  & 0.452      & 0.714 & 0.647  & 0.583\\
Gibert et al.~\cite{Gibert}                      & 0.623   & 0.415  & 0.557            & 0.750     & \textbf{0.578}     & 0.653    & 0.578  & 0.441      & 0.690 & *      & 0.587  \\
Salmenin et al.~\cite{salminen}                    & 0.751       & 0.580  & 0.610            & *         & 0.467     & 0.723    & 0.477  & 0.445      & 0.723 & 0.618  & 0.599 \\
Vidgen et al A.~\cite{Vidgenetal21a}                      & 0.762   & 0.516  & 0.590            & 0.596     & 0.443     & 0.692    & *      & 0.419      & 0.704 & 0.718  & 0.605 \\
Vidgen et al. B~\cite{vidgenetal21b}              & 0.782   & 0.653  & *             & 0.808     & 0.389     & 0.541    & \textbf{0.583}  & 0.571      & \textbf{0.756} & \textbf{0.752}  & 0.649 \\
Qian et al.~\cite{Qian}                        & \textbf{0.821}   & \textbf{0.661} & \textbf{0.678}            & \textbf{0.820}     & 0.474     & 0.595    & 0.579  & 0.548      & *     & 0.724  & \textbf{0.656} \\

\midrule
Mean	    &      \textbf{0.662}	& 0.521	& 0.559	& 0.651	& 0.455	& 0.603	& 0.515	& 0.506 & 	0.621 & 0.594  \\
\midrule
\end{tabular}
}
\end{table*}

\begin{table}[t!]
\centering
\caption{System performance: generalised classifier ranking based on overall mean weighted F1-score, list in ascending order by Mean score}
\resizebox{0.45\textwidth}{!}{
\begin{tabular}{lll}
     \toprule
        \textbf{Model}                           & Weighted F1-score    & Rank \\  \midrule
        Qian et al.  \cite{Qian}                 & 0.656                            & 1 \\ %\
        Vidgen et al. B~\cite{vidgenetal21b}     & 0.649                           & 2 \\ %\
        Vidgen et al. A~\cite{Vidgenetal21a}     & 0.605                            & 3 \\ %\
        Salminen et al.~\cite{salminen}          & 0.599                           & 4 \\
        Gibert et al.~\cite{Gibert}              & 0.587                           & 5 \\ 
        Davidson et al.~\cite{Davidson}          & 0.583                           & 6 \\ %\
        Kennedy et al.~\cite{Kennedy}            & 0.581                           & 7 \\ %\
        Suryawanshi et al.~\cite{Suryawanshi}    & 0.531                           & 8 \\
        Waseem and Havoy~\cite{waseemhovy2016hateful}           & 0.532                           & 9 \\ %\
        Gomez et al.~\cite{Gomez}                & 0.363                           & 10 \\ %\
    \end{tabular}
}
    \label{tab:result.generalised.rank}
\end{table}

\subsection{generalised Learning Experiment.}
In this experiment, we  evaluated our baseline classifier by training it on one dataset and testing it on nine other datasets, excluding the dataset used for training. We repeated the experiment for all the datasets. The results from this experiment provide a clear indication of the suitability of each dataset for producing a classifier that can be generalised to unseen data. We refer to the baseline classifier in this experiment as ``generalised classifier''\footnote{Note: this type of evaluation could also} be referred to as ``transfer learning''. 

Table~\ref{tab:generlized.model} shows the performance of the classifier during this experiment. The first column contains the dataset  used for training the classifier. The other columns (column 1 to 10) contain the datasets used for testing the performance of the classifier. An asterisk `*' in each column indicates the classifier is not tested on the dataset specified in that column. For example, the `*'  in the first row of the second column indicates the classifier is trained on Davidson et al.'s~\cite{Davidson} dataset but not tested on that dataset. This is due to the nature of transfer learning method. Thus, there is one `*' in each row. The numbers in the rows represent the performance of each classifier when tested on all the datasets except the one that is used for training.

The database published by~\cite{Qian} is one of the most suitable for producing a generalised deep learning baseline classifier that performs well on multiple hatespeech datasets. When the baseline classifier was trained on Qian et al.~\cite{Qian}'s dataset, it performed well on four out of nine datasets (namely, Davidson et al.~\cite{Davidson}, Waseem and Havoy~\cite{waseemhovy2016hateful}, Vidgen et al. A~\cite{Vidgenetal21a}, and Salmenin et al.~\cite{salminen}). The classifier's performance on this dataset is shown in bold in the 10th row of Table~\ref{tab:generlized.model}.  

Second to Qian et al.~\cite{Qian}' dataset is Vidgen et al. B~\cite{vidgenetal21b}'s dataset, which performed  well on three out of nine datasets (namely, Vidgen et al.~\cite{Vidgenetal21a}, Qian et al.~\cite{Qian}, and Gibert et al.~\cite{Gibert}). The classifier performed well on only one out of nine datasets when trained on the following dataset: Davidson et al.~\cite{Davidson}, Waseem and Havoy~\cite{waseemhovy2016hateful}, and Gibert et al.~\cite{Gibert}. The classifier trained on Salmenin et al.~\cite{salminen}, Gomez et al.~\cite{Gomez}, Kennedy et al.~\cite{Kennedy}, Vidgen et al.~\cite{vidgenetal21b}, and Suryawanshi et al.~\cite{Suryawanshi} did not perform better than those trained on other datasets. However, it should be noted that their mean score across the nine datasets affected their ranking performance.  

Table~\ref{tab:result.generalised.rank} shows the ranking of the generalised classifier performance for each dataset based on the overall mean performance across the nine datasets used for testing the classifier. The average mean score of the classifier indicates the generalization level of the classifier when trained on each dataset for hatespeech binary classification, i.e., the suitability of a dataset for training a baseline deep learning classifier in a generalised learning setting.

 The classifier ranked first when trained on the Qian et al.~\cite{Qian} dataset and tested on nine other datasets,  achieving 0.656. The classifier trained on the Vidgen et al. B~\cite{vidgenetal21b} dataset ranked second with a mean of 0.649\. The classifier performed worst when trained on Gomez et al.~\cite{Gomez} dataset and tested on the other nine datasets, achieving the lowest mean  weighted F1-score of 0.363. This poor performance indicates that the Gomez et al.~\cite{Gomez} dataset is the least suitable for producing a baseline deep learning classifier in a generalised learning setting. 

The other datasets (Vidgen et al. A.~\cite{Vidgenetal21a}, Salminen et al.~\cite{salminen}, Gibert et al.~\cite{Gibert}, Davidson et al.~\cite{Davidson}, Kennedy et al.~\cite{Kennedy}, Suryawanshi et al.~\cite{Suryawanshi} and Waseem and Havoy~\cite{waseemhovy2016hateful})  achieved mean weighted F1-score between 0.532 and 0.599, as shown in Table~\ref{tab:result.generalised.rank}.

\section{Discussion}
The classifier  performed well when trained and tested on a single dataset at a time, which we referred to in Section~\ref{section.evaluation} as mono-dataset experiment. It produced a weighted F1-score between 0.81 and 0.93 for six out of ten datasets that we used for training and testing the classifier. In order to evaluate each datasets, we applied a generalised learning approach in a second experiment, referred to  as ``generalised learning experiment''. In this experiment, we  trained the classifier on one dataset and tested it on nine other datasets. This approach allows us to rigorously examine the suitability of a dataset for training and testing a classifier in a generalised learning setting, which provides a more rigorous evaluation than mono-dataset evaluation.

We found that the classifier's performance varied depending on the dataset used for training. In this section, we will highlight some of the major features of the datasets that have potentially influenced the classifier's performance. Additionally. we will present several confusion matrices to illustrate some of the classification errors the classifier made when applied to different dataset.

\begin{table}[t]
\centering
\caption{Hate/Not-Hate Hate terms t-test}
\resizebox{0.6\textwidth}{!}{
\begin{tabular}{lllll}\toprule
\textbf{Dataset}        & T-value    & P-value   & Rank P-value & Rank by T-value \\
\midrule
Qian et al ~\cite{Qian}                                     & -3.9461     & 0.0001   & 1            & 1              \\
Gibert et al.~\cite{Gibert}             & -3.5345     & 0.0004   & 2            & 2 \\
Kennedy et al~\cite{Kennedy}           & -3.4692     & 0.0005   & 3            & 3              \\
Vidgen et al. B~\cite{vidgenetal21b}    & -3.4119     & 0.0007   & 4            & 4              \\
Vidgen et al. A~\cite{Vidgenetal21a}      & -2.3177     & 0.0205   & 5            & 5              \\
Waseem and Havoy~\cite{waseemhovy2016hateful}          & -1.7845     & 0.0744   & 6            & 6              \\
Salmenin et al.~\cite{salminen}         & -1.7490     & 0.0804   & 7            & 7              \\
Suryawanshi et al.~\cite{Suryawanshi}   & -1.6805     & 0.0930   & 8            & 8              \\
Davidson et al.~\cite{Davidson}         & -1.6402     & 0.1011   & 9            & 9              \\
Gomez et al.~\cite{Gomez}               & 0.0067      & 0.9947   & 10           & 10

\end{tabular}
}
\label{tab:results.ttest}

\end{table}

\subsection{p-Test}
We have calculated two statistical measures ($P-test$ and $T-test$) using the content of each dataset based on the labels  ``Hate'' or ``Not-Hate''. Details of our approach to computing these statistics are presented in section~\ref{subsec.statistical.analysis}. Since machine learning algorithm performance is based on its learning from identifiable patterns in a given dataset, the $p$ value offers good indication on the available patterns in the ten datasets we have chosen for this study. A $p$ value of 0 indicates that the patterns in the dataset occurred by chance, which may reflect poor dataset annotation. A $p$ value of 1 indicates there is no difference in the patterns in the dataset. Our data analyses results are shown in Table~\ref{tab:results.ttest}. 

The result of our analyses highlighted that the dataset from~\cite{Gomez}  has consistently performed poorly in both of the experiments: mono-dataset experiment and generalised learning experiment. From table~\ref{tab:results.ttest} it can be noted that $p$ the value for this dataset is very close to 1 (0.9947), which means there is no recognisable pattern between ``Hate'' and ``Not-Hate'' content in this dataset. The lack of distinguishable patterns in the dataset highlights the main reason for the baseline classifier failing to learn sufficiently from this dataset, hence performing poorly, producing a weighted F1-score of 0.363 in the generalised learning experiment.

In contrast, as can be seen from table~\ref{tab:result.generalised.rank}, the classifier produced a weighted F1-score of more than 0.5 for all those datasets with $p$ value 0.0001 and <0.1011, indicating that the model learned sufficient patterns to produce a weighted F1-score of over 0.531.

\begin{figure*}[t!]
    \centering
     \begin{subfigure}[b]{0.33\textwidth}
         \centering
         \includegraphics[width=\textwidth]{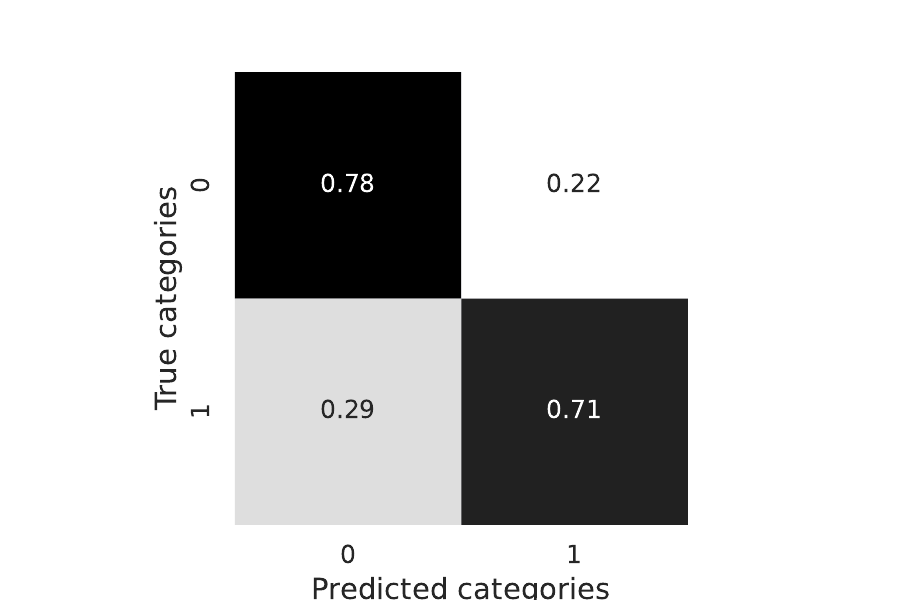}
         \caption{Vidgen et al. B}
         \label{fig:mono.Vidgen et al. B}
     \end{subfigure}
     \hfill
    \begin{subfigure}[b]{0.33\textwidth}
         \centering
         \includegraphics[width=\textwidth]{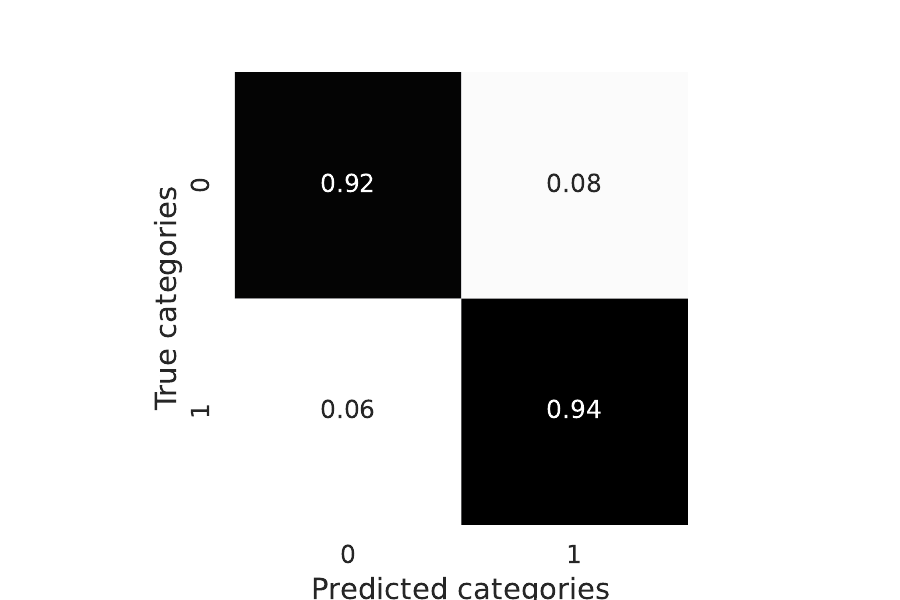}
         \caption{Davidson et al.}
         \label{fig:mono.davidson}
     \end{subfigure}
     \hfill
     \begin{subfigure}[b]{0.3\textwidth}
         \centering
         \includegraphics[width=\textwidth]{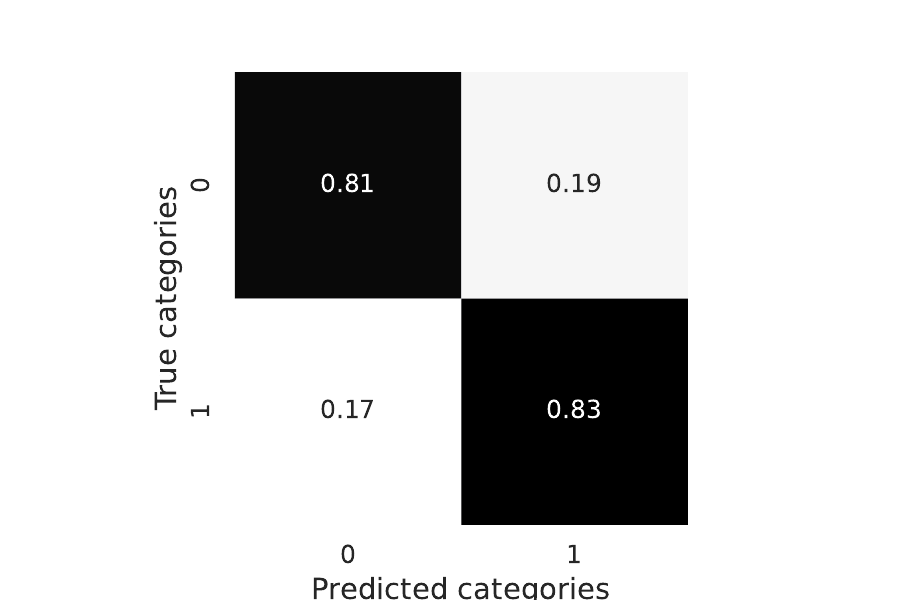}
         \caption{Qian et al.}
         \label{fig:mono.Qian et al.}
     \end{subfigure}
     \hfill
      \begin{subfigure}[b]{0.3\textwidth}
         \centering
         \includegraphics[width=\textwidth]{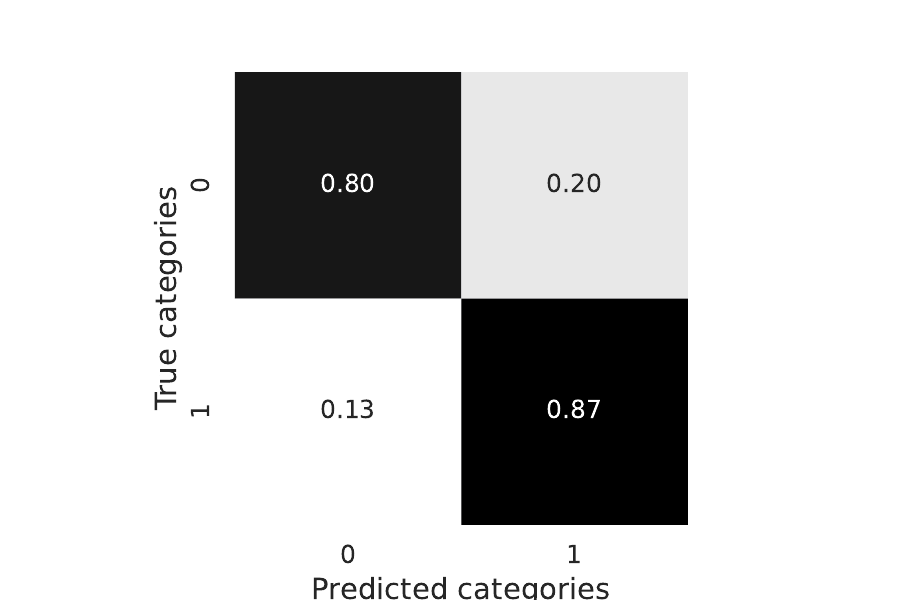}
         \caption{Kennedy et al.}
         \label{fig:mono.kennedyetal}
     \end{subfigure}
     \hfill
     \begin{subfigure}[b]{0.3\textwidth}
         \centering
         \includegraphics[width=\textwidth]{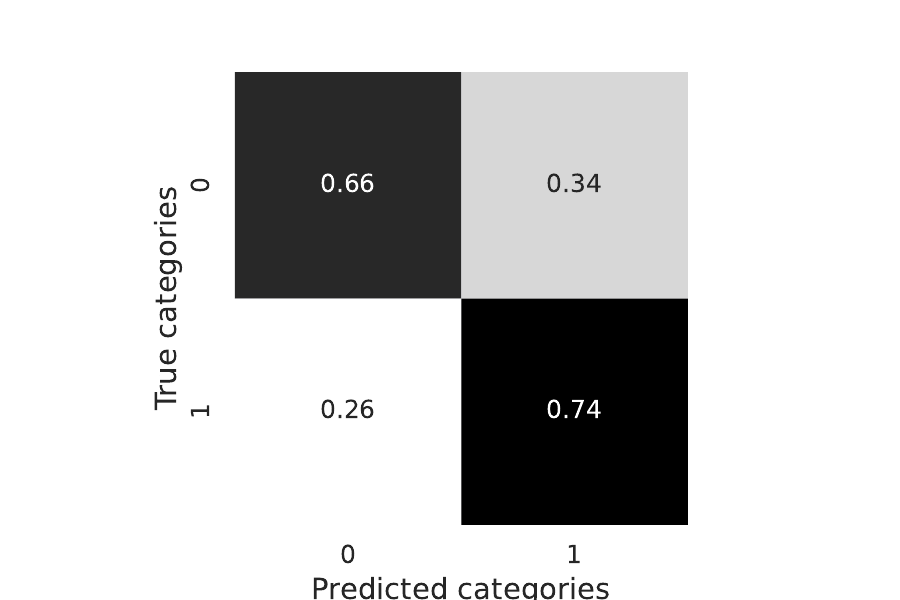}
         \caption{Gomez et al.}
         \label{fig:mono.Gomez et al.}
     \end{subfigure}
     \hfill
     \begin{subfigure}[b]{0.33\textwidth}
         \centering
         \includegraphics[width=\textwidth]{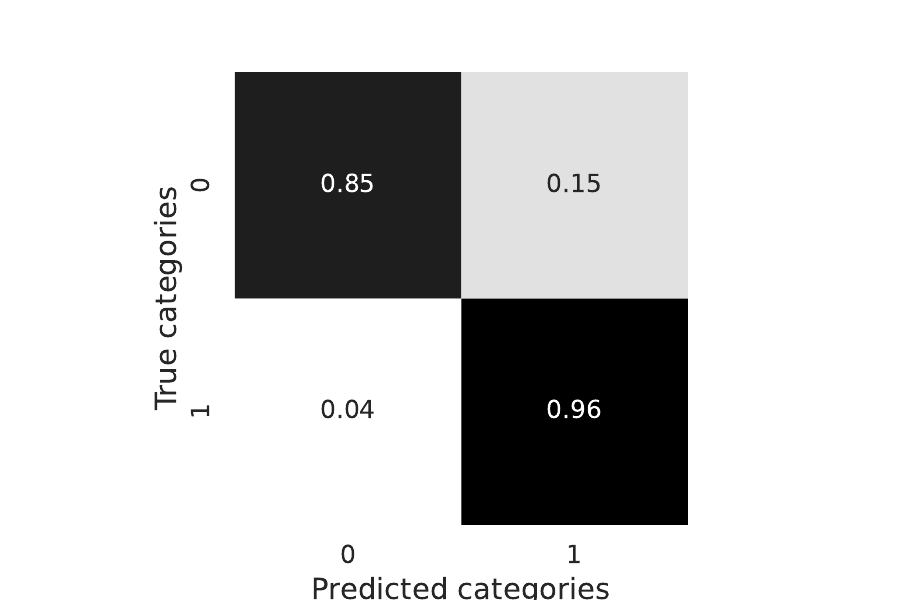}
         \caption{Suryawanshi et al.}
         \label{fig:mono.Suryawanshi et al.}
     \end{subfigure}
     \hfill
     \begin{subfigure}[b]{0.33\textwidth}
         \centering
         \includegraphics[width=\textwidth]{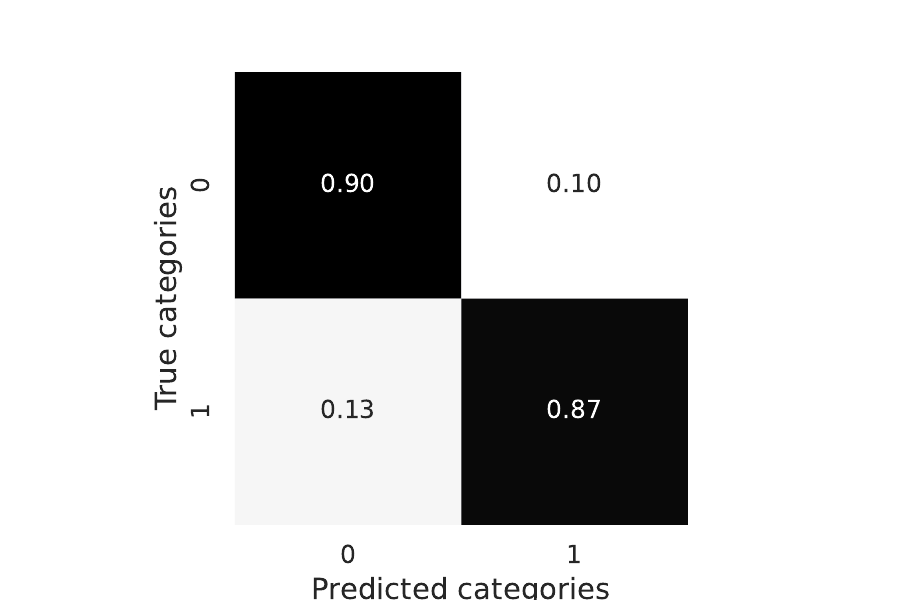}
         \caption{Salmenin et al.}
         \label{fig:mono.salmeninetal}
     \end{subfigure}
     \hfill
     \begin{subfigure}[b]{0.3\textwidth}
         \centering
         \includegraphics[width=\textwidth]{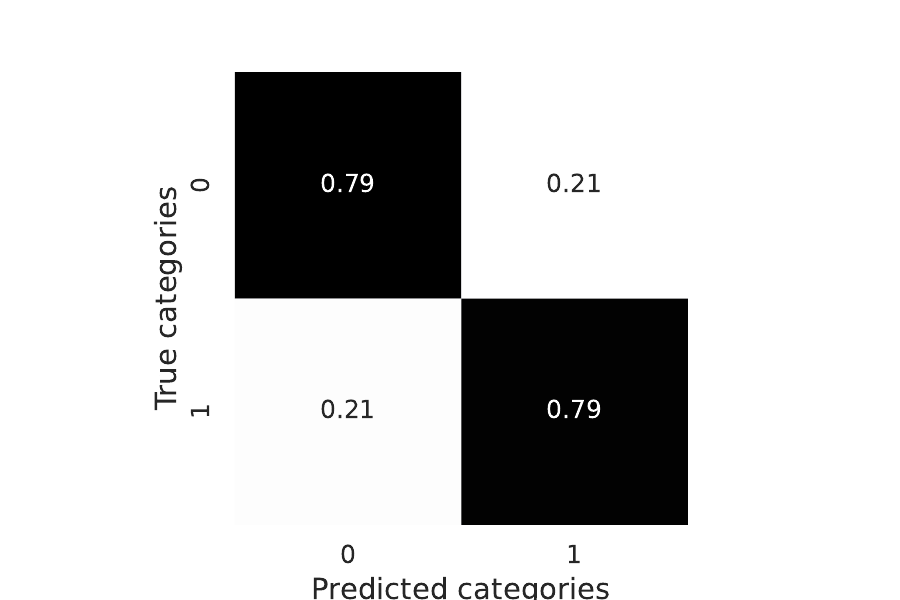}
         \caption{vidgen et al. A}
         \label{fig:mono.vidgenetal}
     \end{subfigure}
     \hfill
     \begin{subfigure}[b]{0.3\textwidth}
         \centering
         \includegraphics[width=\textwidth]{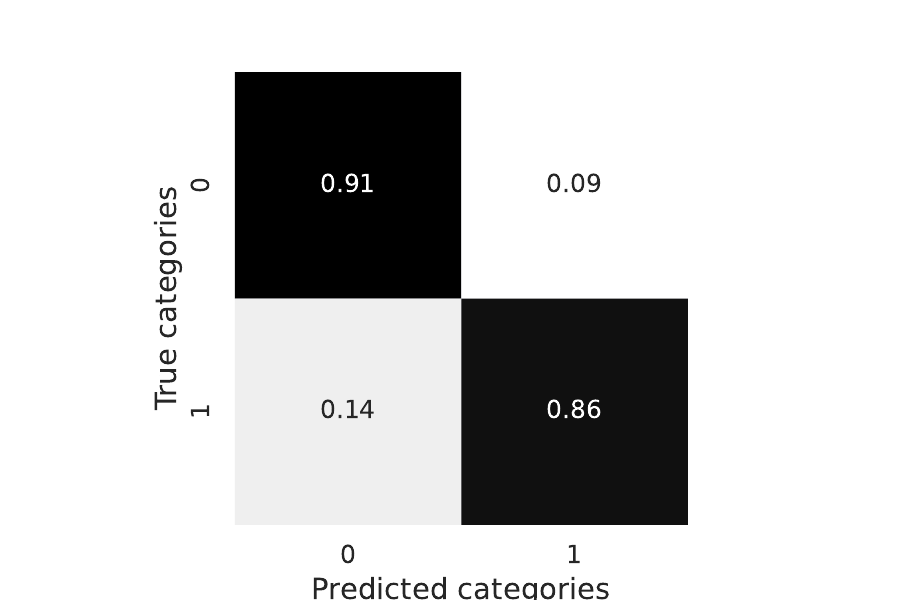}
         \caption{Waseem and Havoy}
         \label{fig:mono.waseem}
     \end{subfigure}
     \caption{Confusion matrix:  mono-dataset classification error analyses}
     \label{fig:cm.mono-dataset}
\end{figure*}

\subsection{Confusion matrix}

In supervised text classification, machine learning algorithms learn from a set of labelled data. Any given labelled dataset contain annotation errors due to many reasons (e.g., annotation procedure, annotator competency, data quality checking, ambiguities in natural language, etc.). Thus, machine learning algorithms are expected to make mistakes since they learn from  annotated data. We use confusion matrices to highlight the classification errors the classifier made in each experiment (mono-dataset and generalised learning experiments). Due to space limitations, we provide comprehensive details on the classification errors of the best and worst performing classifiers compared to other models.

\subsubsection{mono-dataset classifier error analyses}
%Fig  shows the confusing matrix for the mono-dataset experiment for each dataset.

Training the classifier on the dataset published by Davidson et al. ~\cite{Davidson} produced the lowest classification error rate. The classifier correctly classified 92\%  ``Not-Hate' content  and 94\% ``Hate'' content. However, the classifier  miss-classifies 8\% of ``Not-Hate'' content as ``Hate'' and 6 of it ``Hate'' content as ``Not-Hate''. The classifier's misclassification total error rate between the classes (``Hate'' and ``Not-Hate'') is (14\%), as shown in Fig~\ref{fig:mono.davidson}, which is lower compared to when the classifier is trained on  other dataset.

Comparing the confusion matrix in Fig~\ref{fig:cm.mono-dataset}, it appears that the largest misclassification error rate is produced by the classifier when trained on the Gomez et al.~\cite{Gomez} dataset, as shown in  Fig~\ref{fig:mono.Gomez et al.}. The classifier makes a 26\% misclassification error rate for ``Hate'' content and a 34\% error rate  for ``Not-Hate' content.  The classifier seems to perform consistently when trained on Vidgen et al. A's~\cite{Vidgenetal21a} dataset, correctly classifying 79\% of both``Hate'' and ``Not-Hate'' content,  with 21\% error rate for both types of contents. This 21\% misclassification error rate, when training the classifier on Vidgen et al. A's~\cite{Vidgenetal21a} dataset, is the second-largest misclassification error rate  generated by the classifier after the Gomez et al.~\cite{Gomez} trained classifier. The confusion matrix in Fig~\ref{fig:mono.vidgenetal} presents the classification error rate of the classifier when trained on Vidgen et al. A.~\cite{Vidgenetal21a}

The confusion matrices for  Vidgen et al. B (Fig~\ref{fig:mono.Vidgen et al. B}), Salmenine et al. (Fig~\ref{fig:mono.salmeninetal}) and Waseem and Havoy (Fig~\ref{fig:mono.waseem}) show that the classifier performs better  at classifying ``Not-Hate'' content than ``Hate'' content, with a lower misclassification rate  for ``Not-Hate'' content. In contrast, the confusion matrices for Davidson et al. (Fig~\ref{fig:mono.davidson}), Qian et al. (Fig~\ref{fig:mono.Qian et al.}), Kennedy et al. (Fig~\ref{fig:mono.kennedyetal}), Gomez et al.~\cite{Gomez} (Fig~\ref{fig:mono.Gomez et al.}), and Suryawanshi et al. (Fig~\ref{fig:mono.Suryawanshi et al.}) show the model miss-classifies ``Hate'' content less than ``Not-Hate'' content. 

We found that five out of the ten trained classifiers on different dataset, misclassify ``Not-Hate'' content more than ``Hate'' content. Three trained classifiers misclassify  ``Hate'' content more than ``Not-Hate'' content. The exception is the classifier trained on Vidgen et al. A's dataset~\cite{Vidgenetal21a}, which has equal error rates for both ``Hate'' and ``Not-Hate''.

The lowest misclassification rate (4\%) for ``Hate'' content comes from the classifier trained on the Suryawanshi et al. dataset,  indicating that this model  achieves the highest correct classification rate (96\%) for ``Hate'' content. The smallest misclassification error rate  for ``Not-Hate'' comes from the classifier trained on Davidson et al.~\cite{Davidson}, which means that this classifier achieves the highest correct classification rate (92\%) for ``Not-Hate'' content. 

Although the classifier based on the Suryawanshi et al. dataset produces the highest correct classification rate of ``Hate'' content, it  falls behind the classifier based on Davidson et al.~\cite{Davidson} dataset due to its misclassification error rate for ``Not-Hate'' content. in the Suryawanshi et al. -based classifier is nearly twice  as high as that of the Davidson et al.'s dataset-based classifier (15\% vs 8\%).

\begin{figure*}[t!]
    \centering
    \begin{subfigure}[b]{0.3\textwidth}
         \centering
         \includegraphics[width=\textwidth]{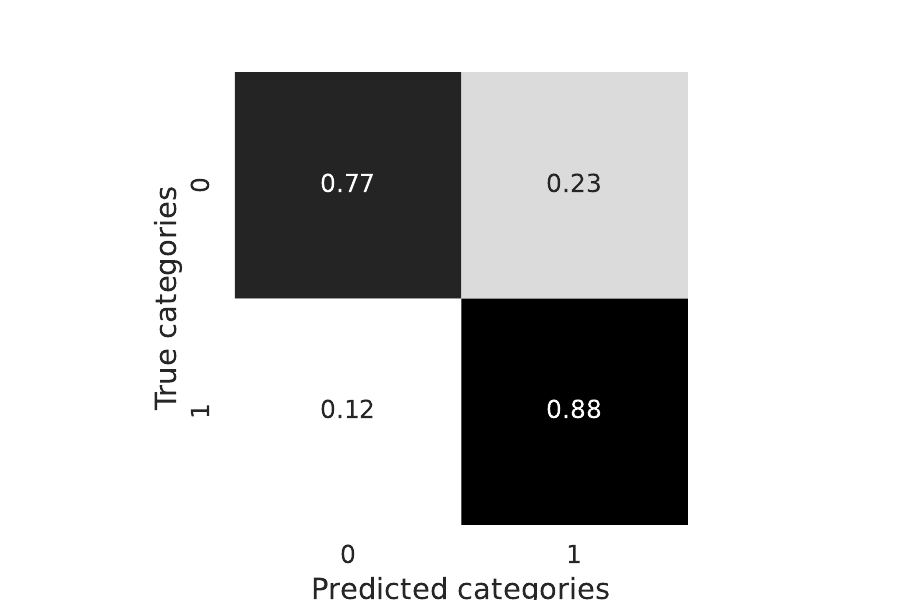}
         \caption{Davidson et al.}
         \label{fig:DavidsonQian et al.}
     \end{subfigure}
     \hfill
     \begin{subfigure}[b]{0.3\textwidth}
         \centering
         \includegraphics[width=\textwidth]{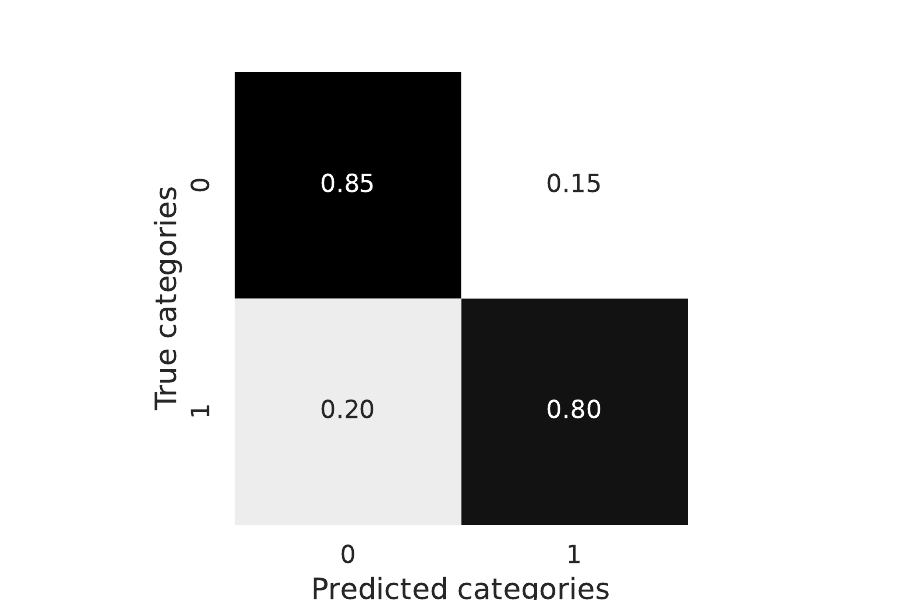}
         \caption{Salmenin et al.}
         \label{fig:SalmeninetalQian et al.}
     \end{subfigure}
     \hfill
      \begin{subfigure}[b]{0.3\textwidth}
         \centering
         \includegraphics[width=\textwidth]{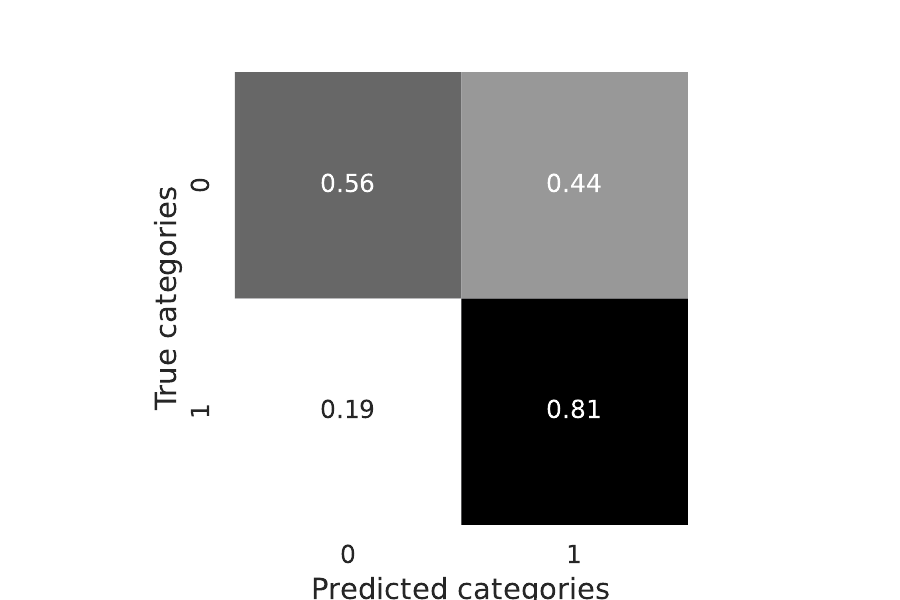}
         \caption{Vidgen et al. B}
         \label{fig:Vidgen et al. BQian et al.}
     \end{subfigure}
     \hfill
    \begin{subfigure}[b]{0.3\textwidth}
         \centering
         \includegraphics[width=\textwidth]{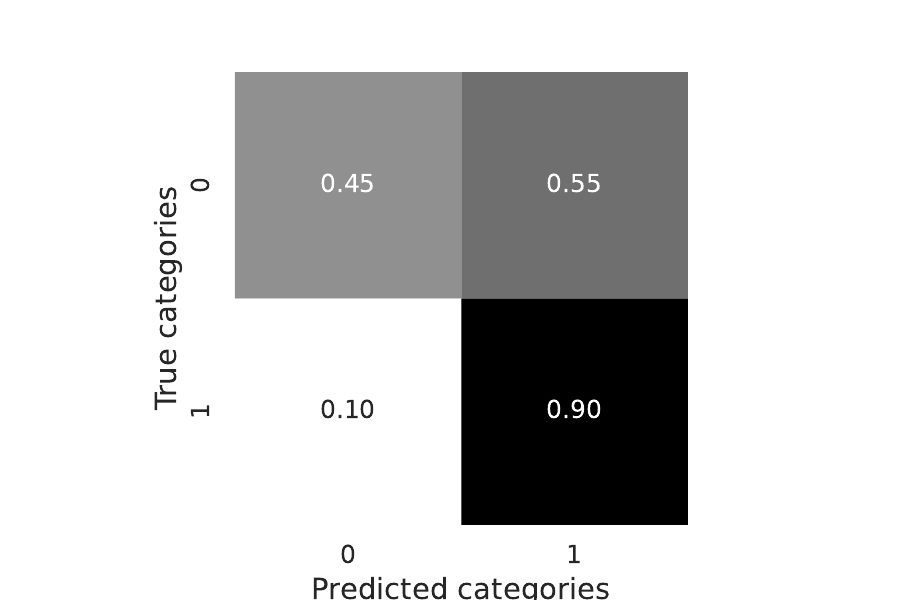}
         \caption{Waseem and Havoy}
         \label{fig:waseemQian et al.}
     \end{subfigure}
     \hfill    
     \begin{subfigure}[b]{0.3\textwidth}
         \centering
         \includegraphics[width=\textwidth]{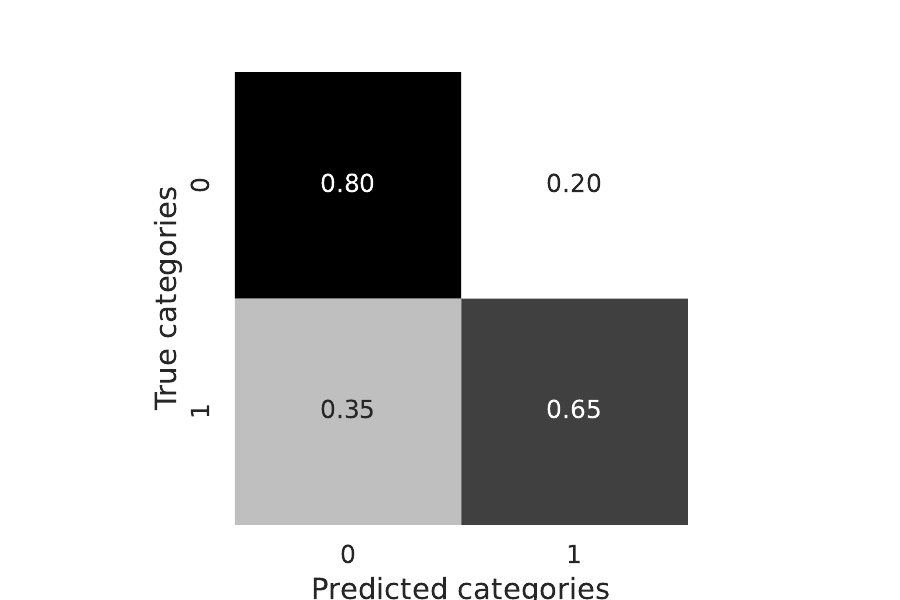}
         \caption{Gibert et al.}
         \label{fig:Gibert et al.Qian et al.}
     \end{subfigure}
     \hfill
     \begin{subfigure}[b]{0.3\textwidth}
         \centering
         \includegraphics[width=\textwidth]{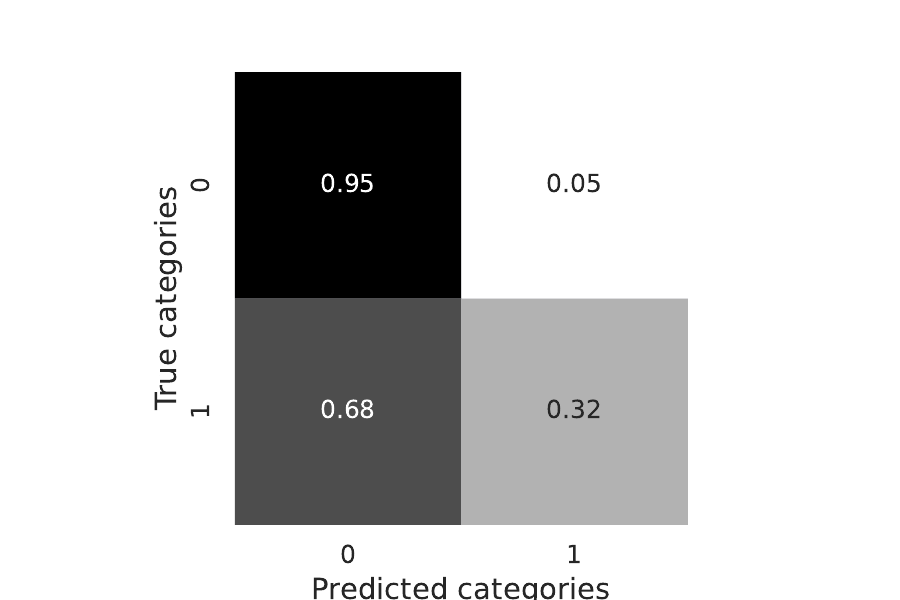}
         \caption{Kennedy et al.}
         \label{fig:KennedyetalQian et al.}
     \end{subfigure}
     \hfill
     \begin{subfigure}[b]{0.3\textwidth}
         \centering
         \includegraphics[width=\textwidth]{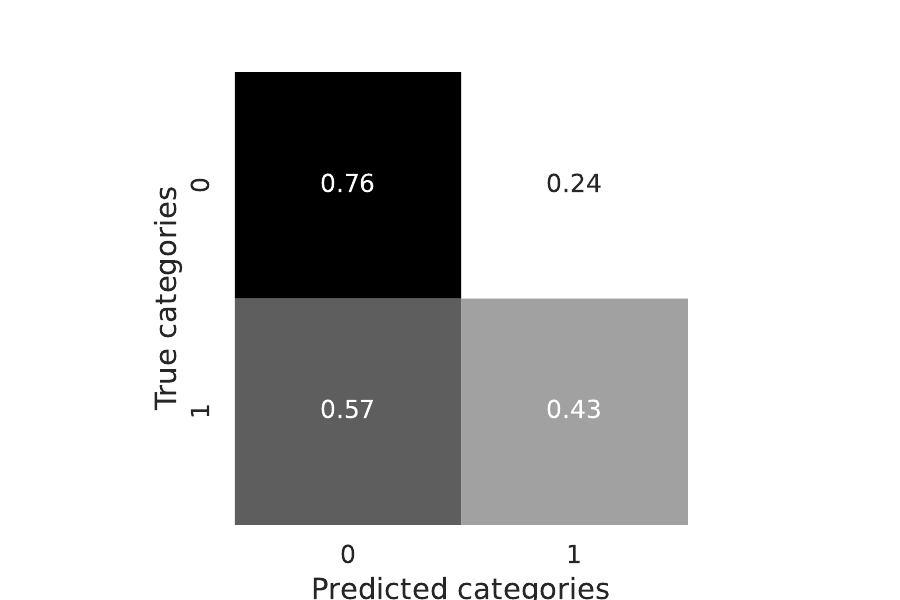}
         \caption{vidgen et al.}
         \label{fig:vidgenetalQian et al.}
     \end{subfigure}
     \hfill
     \begin{subfigure}[b]{0.3\textwidth}
         \centering
         \includegraphics[width=\textwidth]{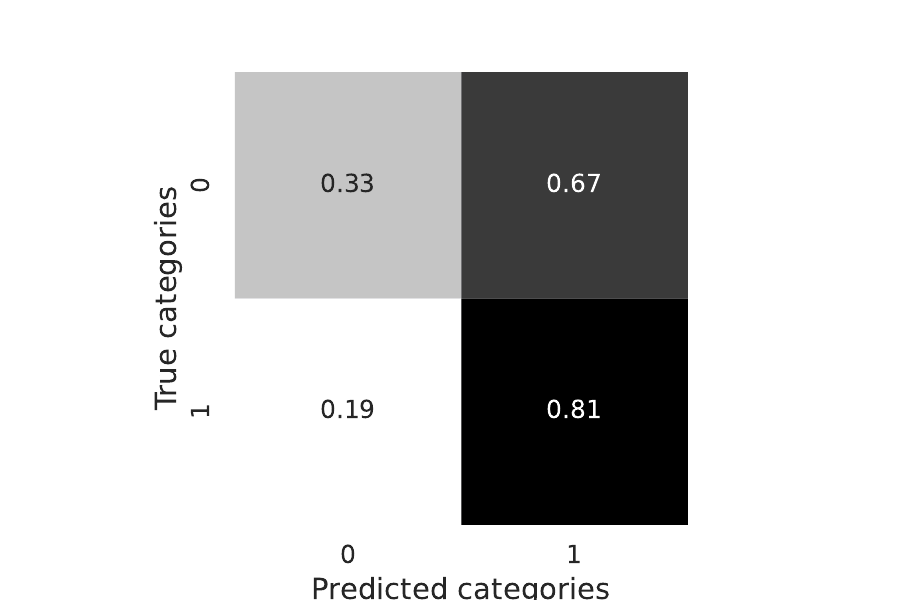}
         \caption{Suryawanshi et al.}
         \label{fig:Suryawanshi et al.Qian et al.}
     \end{subfigure}
     \hfill
     \begin{subfigure}[b]{0.3\textwidth}
         \centering
         \includegraphics[width=\textwidth]{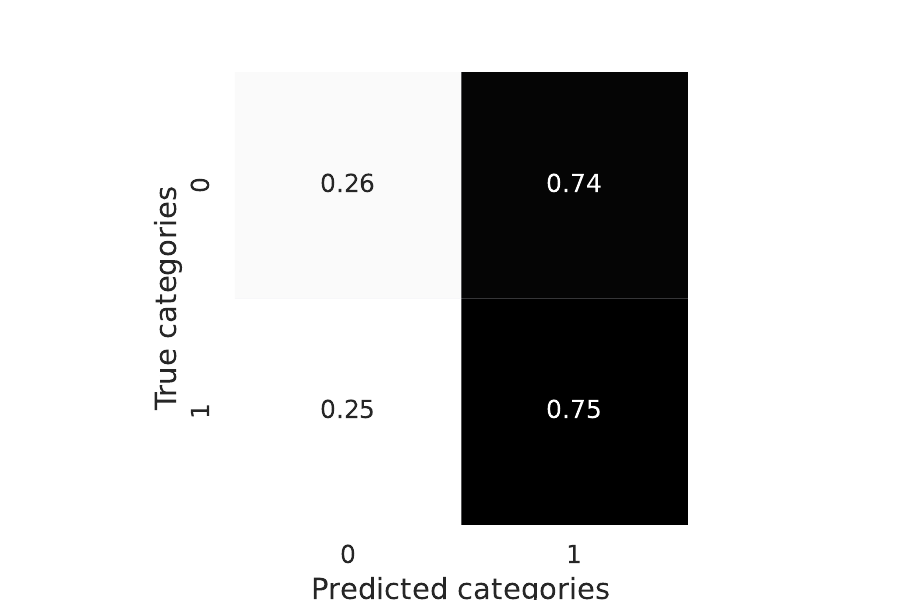}
         \caption{Gomez et al.}
         \label{fig:Gomez et al.Qian et al.}
     \end{subfigure}

     \caption{Confusion matrix: generalised model train on Qian et al. dataset and tested on multiple dataset}
     \label{fig:cm.Qian et al..transfer.learning}
\end{figure*}

\subsubsection{generalised model's error analyses}

In this section, we examine the confusion matrix graphs to analyse the errors made by each classifier when evaluated on transfer learning performance. 

In the generalised learning experiment, we  trained our classifier on one dataset and tested it on the remaining nine datasets. This experiment produced ten classifiers, each of which tested on nine datasets, excluding the one was used for training. For each experiment, we obtained one confusion matrix, resulting in nine confusion matrices per classifier. Therefore, this substantial experiment produced a total of ninety confusion matrices. Due to space limitations, we will focus our discussion only on  a subset of confusion matrices. Specifically, we will examine the classification errors of  the classifiers that have the highest or the lowest mean score of weighted F1-score when tested on the nine different datasets. As shown in Table~\ref{tab:result.generalised.rank}, the highest weighted F1-score (0.656) produced by the classifier trained on the Qian et al.~\cite{Qian}'s dataset, while  the lowest weighted F1-score (0.636) was produced by the classifier trained on Gomez et. al.~\cite{Gomez}'s dataset.

\begin{figure*}[t!]
    \centering
     \begin{subfigure}[b]{0.3\textwidth}
         \centering
         \includegraphics[width=\textwidth]{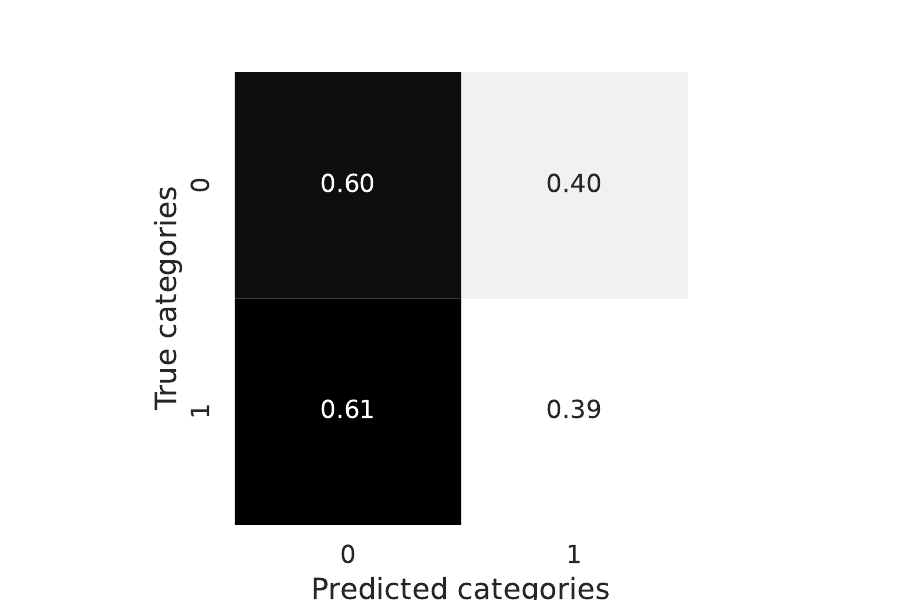}
         \caption{Suryawanshi et al.}
         \label{fig:Suryawanshi et al.Gomez et al.}
     \end{subfigure}
     \hfill
     \begin{subfigure}[b]{0.3\textwidth}
         \centering
         \includegraphics[width=\textwidth]{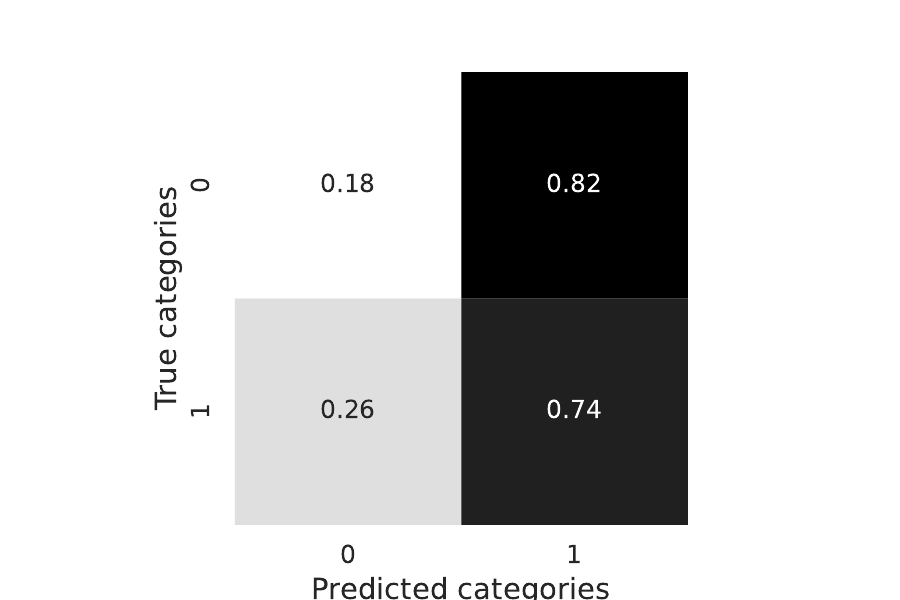}
         \caption{Vidgen et al. A}
         \label{fig:vidgenetalGomez et al.}
     \end{subfigure}
     \hfill
     \begin{subfigure}[b]{0.3\textwidth}
         \centering
         \includegraphics[width=\textwidth]{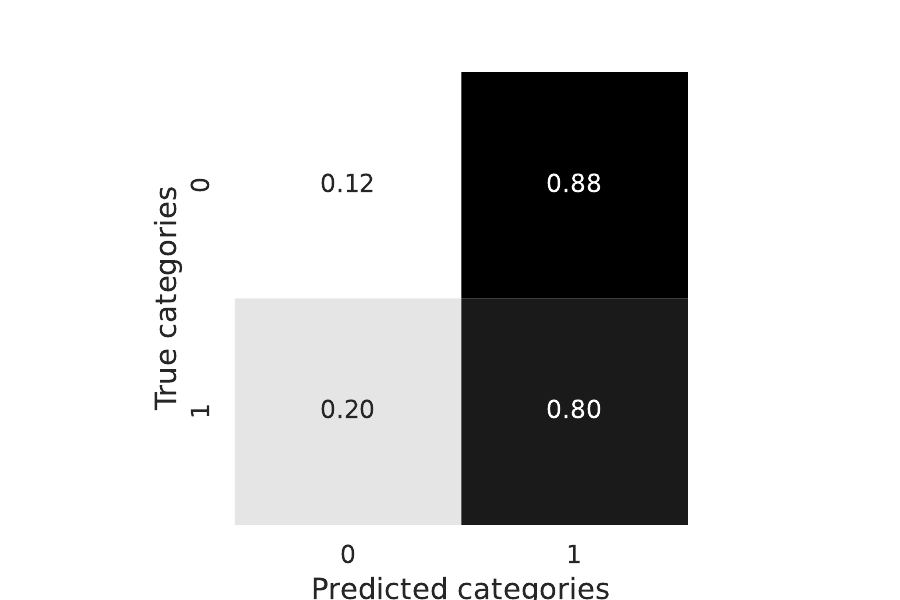}
         \caption{Kennedy et  al.}
         \label{fig:KennedyetalGomez et al.}
     \end{subfigure}
     \hfill
     \begin{subfigure}[b]{0.3\textwidth}
         \centering
         \includegraphics[width=\textwidth]{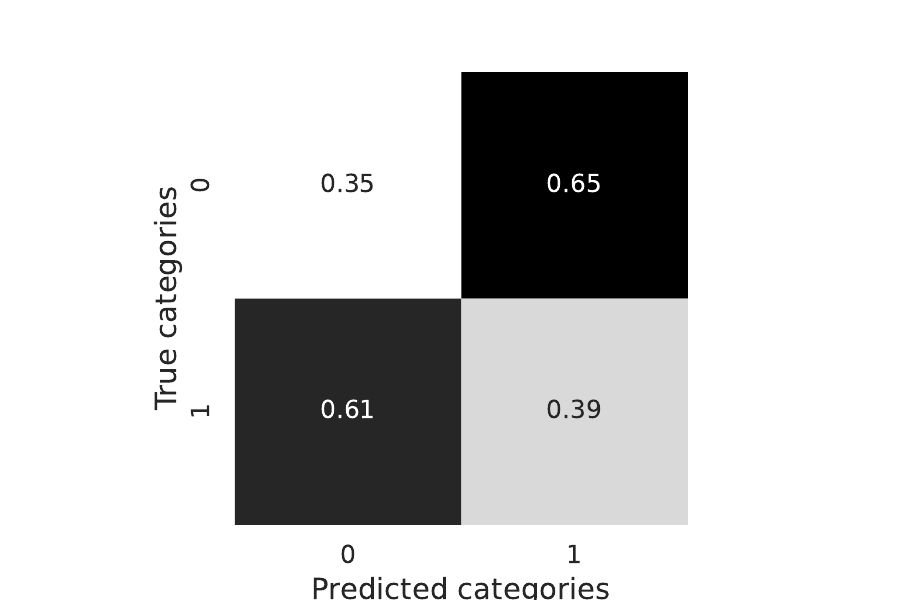}
         \caption{Waseem and Havoy.}
         \label{fig:waseemGomez et al.}
     \end{subfigure}
     \hfill
     \begin{subfigure}[b]{0.3\textwidth}
         \centering
         \includegraphics[width=\textwidth]{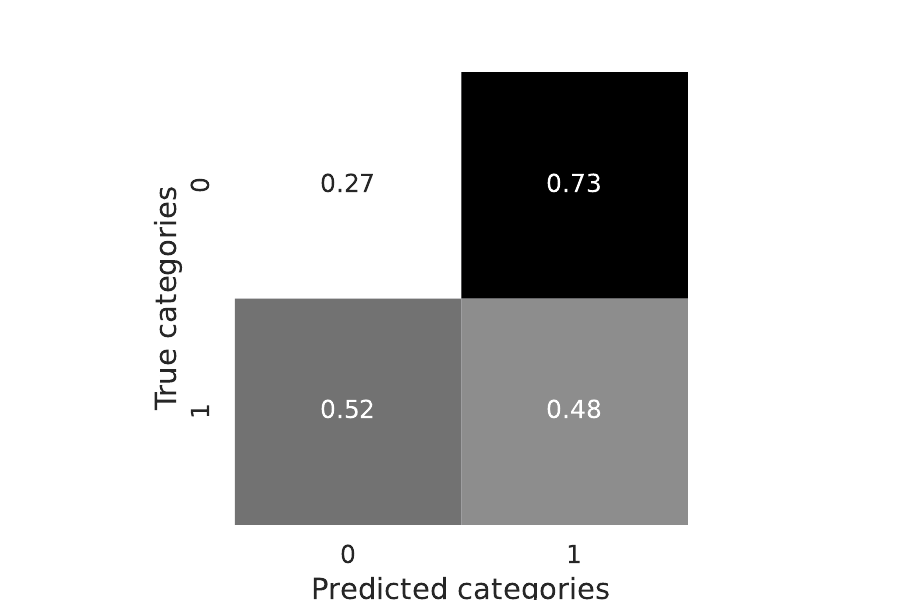}
         \caption{ Vidgen et al. B}
         \label{fig:Vidgen et al. BGomez et al.}
     \end{subfigure}
     \hfill
     \begin{subfigure}[b]{0.3\textwidth}
         \centering
         \includegraphics[width=\textwidth]{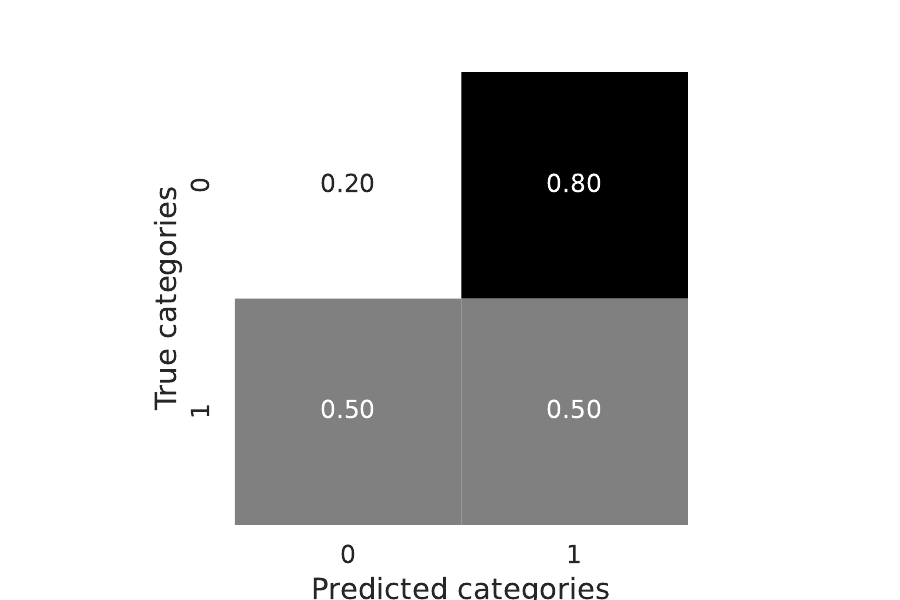}
         \caption{Salmenin et al.}
         \label{fig:SalmeninetalGomez et al.}
     \end{subfigure}
     \hfill
     \begin{subfigure}[b]{0.3\textwidth}
         \centering
         \includegraphics[width=\textwidth]{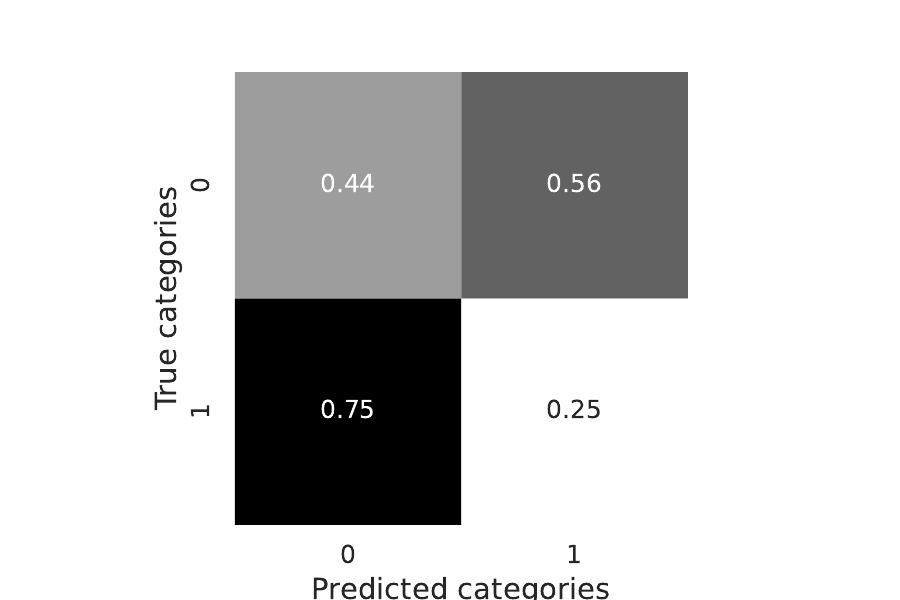}
         \caption{Gibert et al.}
         \label{fig:Gibert et al.Gomez et al.}
     \end{subfigure}
     \hfill
     \begin{subfigure}[b]{0.3\textwidth}
         \centering
         \includegraphics[width=\textwidth]{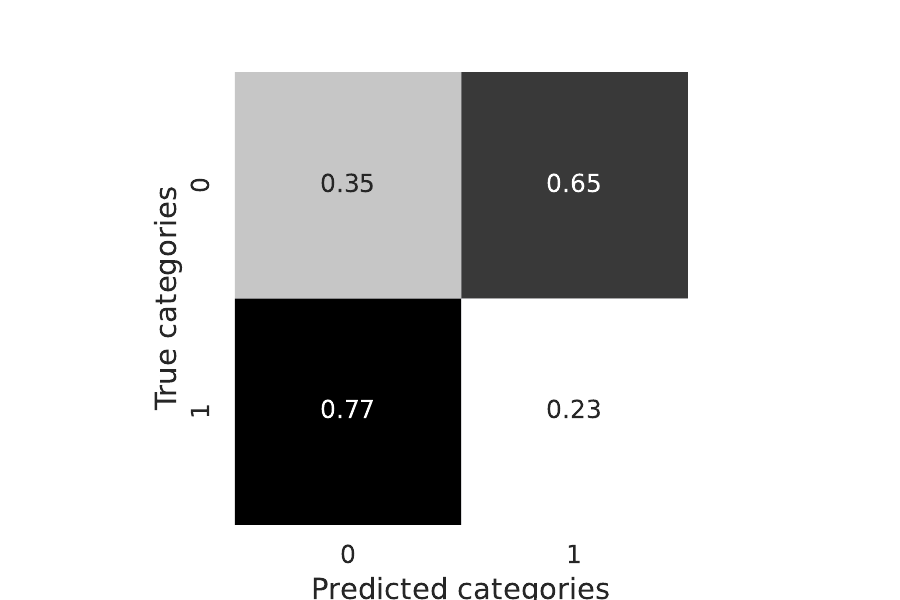}
         \caption{Davidson et al.}
         \label{fig:DavidsonGomez et al.}
     \end{subfigure}
     \hfill     
      \begin{subfigure}[b]{0.3\textwidth}
         \centering
         \includegraphics[width=\textwidth]{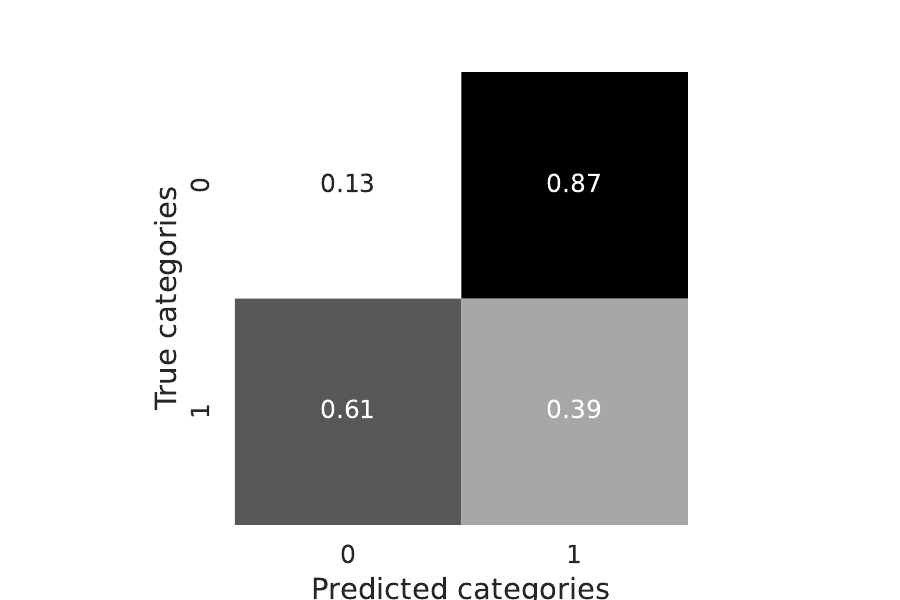}
         \caption{Qian et al.}
         \label{fig:Qian et al.Gomez et al.}
     \end{subfigure}
     \caption{Confusion matrix: generalised classifier trained on Gomez et al. dataset and tested on multiple dataset}
     \label{fig:cm.Gomez et al..transfer.learning}
\end{figure*}

The confusion matrices in Fig~\ref{fig:cm.Qian et al..transfer.learning} and~\ref{fig:cm.Gomez et al..transfer.learning}. are presented in ascending order based on the  weighted F1-score, which is shown in Table~\ref{tab:generlized.model}.  

We grouped our analyses of the confusion matrices based on the following criteria: i) the classifier produced the highest mean weighted F1-score compared to the other nine models, ii) the classifier produced the lowest mean weighted F1-score.

\textbf{generalised model trained on Qian et al. dataset}
The classier trained on the Qian et al.~\cite{Qian} dataset performed the best on four out of nine datasets: Davidson et al.~\cite{Davidson}, Salmenine~\cite{salminen}, Vidgen et al. B~\cite{vidgenetal21b} and Waseem and Havoy~\cite{waseemhovy2016hateful} (~\ref{fig:DavidsonQian et al.}\-~\ref{fig:waseemQian et al.}).  In these cases, the classier  correctly classified ``Hate'' content more accurately than ``Not-Hate'' content, with the exception of the Salminen et al.~\cite{salminen} dataset, where that classier classified ``Not-Hate'' more accurately than ``Hate'' by a margin of 5\%. 

Although the classifier didn't produce a high weighted F1-score when tested on the Kennedy et al.~\cite{Kennedy} dataset, it appears to perform well in correctly classifying ``Not-Hate'' content. As shown in Fig~\ref{fig:KennedyetalQian et al.}, the classifier has misclassification error rate of just 5\% for ``Not-Hate'' content. However, the classifier has large misclassification error rate 68\% for ``Hate'' content, which is the main reason the classier didn't produce high weighted F1-score on this dataset compared to the performance  on other dataset.

In contrast, the classifier's large misclassification error rate of 74\% for ``Not-Hate'' content also negatively impacts the classifier's weighted F1-score. The Suryawanshi et al.~\cite{Suryawanshi}. dataset seem to come second after Gomez et al.~\cite{Gomez} dataset in misclassifying ``Not-Hate'' content, with an error rate of 67\%.

\textbf{generalised model trained on Gomez et al. dataset}
Fig~\ref{fig:cm.Gomez et al..transfer.learning} shows the confusing matrix for the classifier  trained on Gomez et al.~\cite{Gomez} dataset.

The classifier trained on the Gomez et al.~\cite{Gomez}. dataset  demonstrated the worst performance compared to all the other classifiers. We refer to this classifier as ``Gomez et al. classifier''. The confusion matrix showing the errors made by this classifier  when tested on nine datasets is shown in Fig~\ref{fig:cm.Gomez et al..transfer.learning}.

 The classifier  higher misclassification error rate on ``not-Hate'' content than ``Hate'' content,  with 61\% of ``Not-Hate'' being classified as ``Hate'', as shown in the Confusion matrix in Fig~\ref{fig:Suryawanshi et al.Gomez et al.}. In contrast, the classifier makes  significant misclassification errors for ``Hate''  content, as presented in the confusion matrices in Fig~\ref{fig:vidgenetalGomez et al.} and~\ref{fig:KennedyetalGomez et al.}, where the misclassification error rate of ``Hate'' content  exceed 80\%. 
 
 For several other datasets, the classifier seems to  struggle to correctly classify either types of content (``Hate'' and ''Not-Hate''). For the Waseem and Havoy~\cite{waseemhovy2016hateful} dataset, the correct classification rate does not exceed 35\%, as shown in Fig~\ref{fig:waseemGomez et al.}. For the Gibert et al.~\cite{Gibert}, Davidson et al.~\cite{Davidson} and Qian et al.~\cite{Qian} datasets, the classifier has a misclassification error rate between 56\% to 87\%  for either ``Hate'' or ``Not-Hate'' content. The exception  is the Salmenine et al.~\cite{salminen} dataset where the classifier make 50/50 misclassification error of ``Not-Hate'' content, but a large error rate of 80\% for ``Hate'' content.

\begin{figure}[t]
    \centering
    \includegraphics[width=0.55\linewidth]{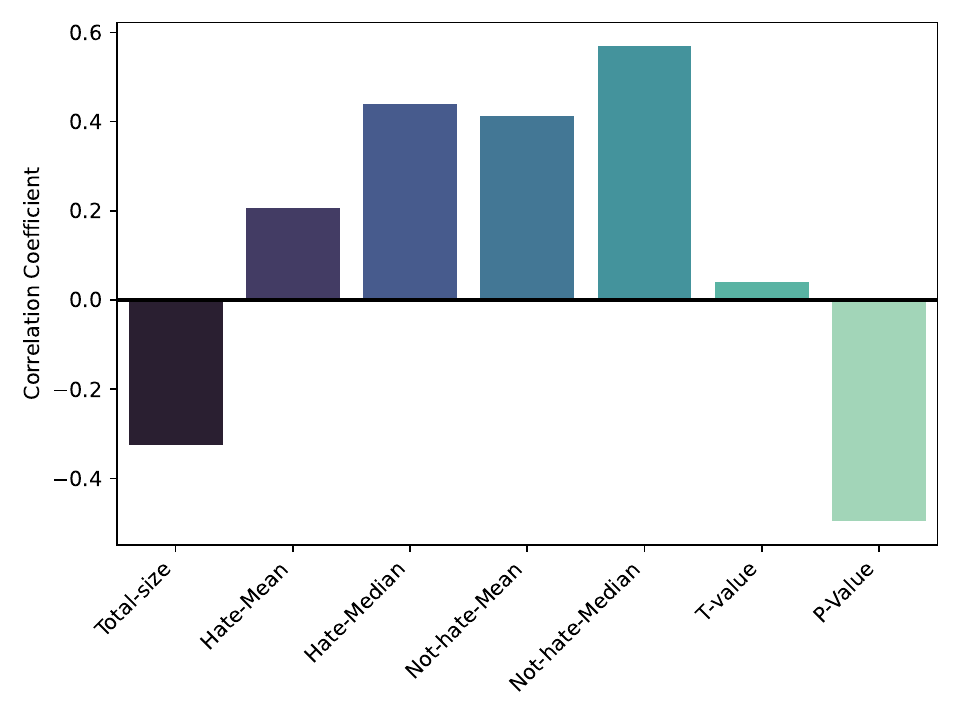}
    \caption{Pearson correlation between each feature and the F1 score of the hatespeech classifier for the mono dataset}
    \label{fig:corr}
\end{figure}

\begin{table}[]
\caption{The regression model predicted F1 score and the actual achieved scores}
\centering
\resizebox{0.3\textwidth}{!}{%
\begin{tabular}{@{}lll@{}}
\toprule
\textbf{Dataset}    & \textbf{LR} & \textbf{F1-score} \\ \midrule
Gomez et al.        & 0.700    & 0.697             \\
Vidgen et al. B    & 0.724    & 0.740             \\
Gibert et al.       & 0.816    & 0.777             \\ 
Vidgen              & 0.847    & 0.789             \\
Qian et al.         & 0.800    & 0.816             \\
Kennedy             & 0.817    & 0.840             \\
Waseem and Havoy    & 0.877    & 0.879             \\
Salminen            & 0.883    & 0.884             \\
Suryawanshi et al.  & 0.904    & 0.902             \\
Davidson            & 0.886    & 0.930        \\  \bottomrule
\end{tabular}
}
\label{tab:pred_values}
\end{table}

\subsubsection{Model performance Analysis}
To develop an efficient hatespeech classifier, it is crucial to train the model on a high-quality dataset that provides sufficient information for accurate classification and real-world implementation. To gain a deeper understanding of the role of features in the effectiveness of the classifier, we conducted a correlation test between the dataset features and the classification F1 score. This analysis helps to identify the most informative features and optimize the dataset for classifier performance. Fig \ref{fig:corr} presents the results of our correlation analysis for the mono dataset test, showing the Pearson correlation between each feature and the F1 score of the classifier. This information can be used to identify the most informative features of the dataset.

Based on our analysis, we observed that the Median word count for the not-hate part of the datasets and the P-value are highly correlated, indicating that these features have a significant impact on the performance of the hatespeech classifier. On the other hand, the T-value is the least correlated feature, suggesting that it may not have a significant impact on the classifier's effectiveness.

This finding highlights the importance of statistical characteristics of the dataset used in the hatespeech classifier, as some characteristics may have a more significant impact on the model's performance than others. By understanding the correlation between the characteristic and the model's effectiveness, we can optimize the dataset balancing and improve the accuracy of the classifier.
 
To ensure the validity of our correlation test, we implemented a linear regression model that was trained on the statistical characteristics of the dataset and used to predict the F1 score of the hatespeech classifier. Table \ref{tab:model.performance.prediction} shows the features used in the regression model.
The regression model produced accurate predictions of the classification model, with a coefficient of determination of 0.84. This high level of accuracy demonstrates the reliability of our correlation test and supports the conclusion that the identified features have a significant impact on the effectiveness of the classifier. The predicted values and the achieved F1-scores are presented in Table \ref{tab:pred_values}.

\begin{figure}[t]
    \centering
    \includegraphics[width=0.5\linewidth]{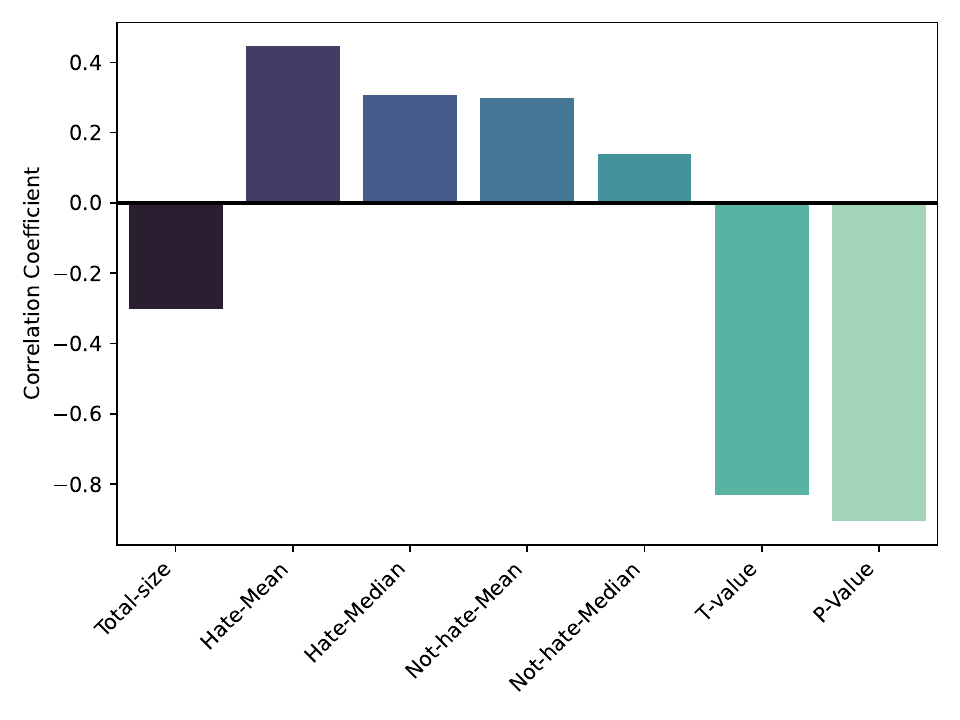}
    \caption{Pearson correlation between each feature and the F1 score of the hatespeech classifier for the Generalised Learning Experiment.}
    \label{fig:corr_gen}
\end{figure}

\begin{table}[t]
\centering
\caption{The features for the ML model}
  \resizebox{0.75\columnwidth}{!}{%

\begin{tabular}{@{}lllllllll@{}}
\toprule
               & \multicolumn{7}{l}{Features (X)}                                                                                            & Target (y) \\ \midrule
               &            & \multicolumn{2}{l}{Hate Word Count} & \multicolumn{2}{l}{Not Hate Word Count} & \multicolumn{2}{l}{Hate Terms} & Mono \\
Model          & Total-size & Mean            & Median            & Mean              & Median              & T-value        & P-Value       & F1         \\ \midrule
 Vidgen et al. B & 8186       & 39.5            & 19                & 27.2              & 14                  & -3.4119        & 0.0007        & 0.740 \\ %0.649      \\
Davidson        & 2860       & 13.9            & 13                & 17.8              & 15                  & -1.6402        & 0.1011        & 0.930 \\ %0.583      \\
Qian et al.            & 4696       & 27.8            & 23                & 19.9              & 15                  & -3.9461        & 0.0001        & 0.816 \\ %0.656      \\
Kennedy        & 92042      & 25.8            & 19                & 28                & 21                  & -3.4692        & 0.0005        & 0.840 \\ %0.581      \\
Gomez et al.       & 50526      & 11.7            & 11                & 11.4              & 11                  & 0.0067         & 0.9947        & 0.697 \\ %0.363      \\
Vidgen         & 37938      & 23.8            & 15                & 25.1              & 17                  & -2.3177        & 0.0205        & 0.789 \\ %0.605      \\
Waseem and Havoy         & 5384       & 16.9            & 17                & 14                & 14                  & -1.7845        & 0.0744        & 0.879 \\ %0.532      \\
Suryawanshi et al.       & 606        & 45              & 33                & 44.8              & 32                  & -1.6805        & 0.093         & 0.902 \\ %0.531      \\
Salminen       & 1716       & 43.6            & 34                & 38.2              & 30                  & -1.749         & 0.0804        & 0.884 \\ %0.599      \\
Gibert et al.         & 2874       & 22              & 20                & 17.3              & 15                  & -3.5345        & 0.0004        & 0.777 \\ %0.587      \\ \bottomrule
\end{tabular}
}
\label{tab:model.performance.prediction}
\end{table}

To optimize the hatespeech classifier, we conducted an analysis using the average F1-score in the generalised learning Experiment as the metric for evaluating the classifier's performance. Our analysis revealed that the P-value is still one of the most highly correlated dataset characteristics for the classifier's effectiveness, as shown in Fig \ref{fig:corr_gen}.

However, we also observed that the Median Not-Hate word count, which was previously highly correlated in our initial analysis, has now dropped to become the least correlated characteristic of the training dataset in the generalised Learning Experiment. 
This finding suggests that the importance of certain dataset characteristics may vary depending on the specific experimental conditions, highlighting the importance of comprehensive analysis to identify the optimal dataset for training the hatespeech classifier.

By using these insights to fine-tune the dataset, we can improve the accuracy of the hatespeech classifier in real-world applications, making it a more effective tool for identifying and combating hatespeech.

\section{Conclusions and Future Work}
\label{section.conclusion}
Despite myriad benefits of social network platforms for users and businesses, malicious users abuse them by targeting specific users based on their identity. This phenomenon is referred to as Hatespeech, where malicious users target vulnerable people with abusive text, or graphics, to degrade and cause them harm. The severity of hatespeech on victims has forced researcher, social media platforms and governments to take action to eliminate it. The task of eliminating hatespeech require automation, since manual intervention is time-consuming and expensive for humans to perform. Thus, large numbers of annotated dataset are developed for training machine learning algorithms to automate the detection and classification of hatespeech. However, the available datasets have been annotated with different types of hatespeech in a number of different ways. This introduces various constraints that affect the training of machine learning algorithms on them.
In this study, we have conducted extensive empirical evaluation of multiple hatespeech dataset to examine their suitability for training machine learning algorithms for the task of automated hatespeech classification. Our main contributions  are as followings: (i) we present the first extensive empirical evaluation of many hatespeech dataset, (ii) We empirically demonstrate that the quality of dataset content has a greater positive impact on AI hatespeech classification than factors such as content volume, context diversity, and data modalities, iii) a novel approach to extract and use statistical features from hatespeech dataset and use certain machine learning algorithms to predict deep learning algorithms performance on hatespeech classification, and (iv) offer a baseline deep learning architecture for automated hatespeech classification. 

To complement the dataset evaluation, we have conducted statistical data analyses on the dataset to examine their features. Moreover, we have analysed the system output to highlight and compare systems' error rates and error types based on each dataset.

We have identified several future works. The binary classification of hatespeech could be the first step in automated hatespeech processing. In this project, we  demonstrated the strengths and weaknesses of several public hatespeech datasets  using a binary classification approach. Multi-label classification, which is helpful for identifying specific types of hatespeech, would be helpful in tasks that require identifying specific hatespeech content, such as hatespeech based on race, gender, sexuality, religion etc. Most  public datasets contain different types of hatespeech, but they lack consistency. A second area for future research involves enhancing the content quality of datasets that the baseline classifier struggles to learn from effectively. Our plan is to explore automated methods for relabelling the content of selected datasets. Additionally, in our future work, we intend to evaluate datasets that encompass a common set of hatespeech types. By addressing these aspects, we aim to improve the overall performance and robustness of hatespeech classification models.

\bibliographystyle{unsrt} 
\bibliography{manuscript}

\begin{thebibliography}{10}

\bibitem{Soto2018DataQC}
Axel~J. Soto, Cynthia Ryan, Fernando~Pe{\~n}a Silva, Tapajyoti Das, Jacek Wolkowicz, Evangelos~E. Milios, and Stephen Brooks.
\newblock Data quality challenges in twitter content analysis for informing policy making in health care.
\newblock In {\em Hawaii International Conference on System Sciences}, 2018.

\bibitem{Kennedy}
Chris~J. Kennedy, Geoff Bacon, Alexander Sahn, and Claudia von Vacano.
\newblock Constructing interval variables via faceted rasch measurement and multitask deep learning: a hate speech application, 2020.

\bibitem{Gomez}
Raul Gomez, Jaume Gibert, Llu{\'{\i}}s G{\'{o}}mez, and Dimosthenis Karatzas.
\newblock Exploring hate speech detection in multimodal publications, 2019.

\bibitem{Djuricetal2015}
Nemanja Djuric, Jing Zhou, Robin Morris, Mihajlo Grbovic, Vladan Radosavljevic, and Narayan Bhamidipati.
\newblock Hate speech detection with comment embeddings.
\newblock In {\em Proceedings of the 24th International Conference on World Wide Web}, WWW '15 Companion, page 29–30, New York, NY, USA, 2015. Association for Computing Machinery.

\bibitem{Davidson}
Thomas Davidson, Dana Warmsley, Michael Macy, and Ingmar Weber.
\newblock Automated hate speech detection and the problem of offensive language.
\newblock {\em Proceedings of the International AAAI Conference on Web and Social Media}, 11(1):512--515, May 2017.

\bibitem{salminen}
Joni Salminen, Hind Almerekhi, Milica Milenkovi{\'c}, Soon-gyo Jung, Jisun An, Haewoon Kwak, and Bernard~J Jansen.
\newblock Anatomy of online hate: developing a taxonomy and machine learning models for identifying and classifying hate in online news media.
\newblock In {\em Twelfth International AAAI Conference on Web and Social Media}, 2018.

\bibitem{dinaka212}
Dinakar Karthik, Jones Birago, Havasi Catherine, Lieberman Henry, and Picard Rosalind.
\newblock Common sense reasoning for detection, prevention, and mitigation of cyberbullying.
\newblock {\em ACM Trans. Interact. Intell. Syst.}, 2:11--30, 2012.

\bibitem{Suryawanshi}
Shardul Suryawanshi, Bharathi~Raja Chakravarthi, Mihael Arcan, and Paul Buitelaar.
\newblock Multimodal meme dataset (multioff) for identifying offensive content in image and text.
\newblock In {\em Proceedings of the second workshop on trolling, aggression and cyberbullying}, pages 32--41, 2020.

\bibitem{waseemhovy2016hateful}
Zeerak Waseem and Dirk Hovy.
\newblock Hateful symbols or hateful people? predictive features for hate speech detection on {T}witter.
\newblock In {\em Proceedings of the {NAACL} Student Research Workshop}, pages 88--93, San Diego, California, jun 2016. Association for Computational Linguistics.

\bibitem{schmidt-wiegand-2017-survey}
Anna Schmidt and Michael Wiegand.
\newblock A survey on hate speech detection using natural language processing.
\newblock In {\em Proceedings of the Fifth International Workshop on Natural Language Processing for Social Media}, pages 1--10, Valencia, Spain, April 2017. Association for Computational Linguistics.

\bibitem{Nobataetal2016}
Chikashi Nobata, Joel Tetreault, Achint Thomas, Yashar Mehdad, and Yi~Chang.
\newblock Abusive language detection in online user content.
\newblock In {\em Proceedings of the 25th International Conference on World Wide Web}, WWW '16, page 145–153, Republic and Canton of Geneva, CHE, 2016. International World Wide Web Conferences Steering Committee.

\bibitem{warner-hirschberg-2012-detecting}
William Warner and Julia Hirschberg.
\newblock Detecting hate speech on the world wide web.
\newblock In {\em Proceedings of the Second Workshop on Language in Social Media}, pages 19--26, Montr{\'e}al, Canada, jun 2012. Association for Computational Linguistics.

\bibitem{vidgenetal21b}
Bertie Vidgen, Dong Nguyen, Helen Margetts, Patricia Rossini, Rebekah Tromble, Kristina Toutanova, Anna Rumshisky, Luke Zettlemoyer, Dilek Hakkani-Tur, Iz~Beltagy, et~al.
\newblock Introducing cad: the contextual abuse dataset.
\newblock In {\em Proceedings of the 2021 Conference of the North American Chapter of the Association for Computational Linguistics: Human Language Technologies}, pages 2289--2303. Association for Computational Linguistics, 2021.

\bibitem{Qian}
Jing Qian, Anna Bethke, Yinyin Liu, Elizabeth Belding, and William~Yang Wang.
\newblock A benchmark dataset for learning to intervene in online hate speech, 2019.

\bibitem{Gibert}
Ona De~Gibert, Naiara Perez, Aitor Garc{\'\i}a-Pablos, and Montse Cuadros.
\newblock Hate speech dataset from a white supremacy forum.
\newblock {\em arXiv preprint arXiv:1809.04444}, 2018.

\bibitem{Vidgenetal21a}
Bertie Vidgen, Tristan Thrush, Zeerak Waseem, and Douwe Kiela.
\newblock Learning from the worst: Dynamically generated datasets to improve online hate detection.
\newblock In {\em Proceedings of the 59th Annual Meeting of the Association for Computational Linguistics and the 11th International Joint Conference on Natural Language Processing (Volume 1: Long Papers)}, pages 1667--1682, Online, August 2021. Association for Computational Linguistics.

\bibitem{MehdadandTetreault16}
Yashar Mehdad and Joel Tetreault.
\newblock Do characters abuse more than words? in 17th annual meeting of the special interest group on discourse and dialogue.
\newblock In {\em Proceedings of the 17th Annual Meeting of the Special Interest Group on Discourse and Dialogue}, page 299–303, Los Angeles, CA, USA, 2016.

\bibitem{Xiangetal15}
Guang Xiang, Bin Fan, Ling Wang, Jason Hong, and Carolyn Rose.
\newblock Detecting offensive tweets via topical feature discovery over a large scale twitter corpus.
\newblock In {\em Proceedings of the 21st ACM International Conference on Information and Knowledge Management}, CIKM '12, page 1980–1984, New York, NY, USA, 2012. Association for Computing Machinery.

\bibitem{Gitari2015ALA}
Gitari Njagi, Dennis, Zuping Zhang, Zhang Zuping, Damien Hanyurwimfura, and Long Jun.
\newblock A lexicon-based approach for hate speech detection.
\newblock {\em International Journal of Multimedia and Ubiquitous Engineering}, 10:215--230, 2015.

\bibitem{scipy}
{SciPy 1.0 Contributors}.
\newblock {SciPy} 1.0: Fundamental algorithms for scientific computing in python.
\newblock {\em Nature Methods}, 17:261--272, 2020.

\bibitem{bert}
Jacob Devlin, Ming-Wei Chang, Kenton Lee, and Kristina Toutanova.
\newblock Bert: Pre-training of deep bidirectional transformers for language understanding.
\newblock {\em arXiv preprint arXiv:1810.04805}, 2018.

\bibitem{devlinetal19}
Jacob Devlin, Ming{-}Wei Chang, Kenton Lee, and Kristina Toutanova.
\newblock {BERT:} pre-training of deep bidirectional transformers for language understanding.
\newblock {\em CoRR}, abs/1810.04805, 2018.

\end{thebibliography}

\end{document}